\documentclass[10pt,journal,compsoc]{IEEEtran}

\usepackage[utf8]{inputenc} % allow utf-8 input
\usepackage[T1]{fontenc}    % use 8-bit T1 fonts
\usepackage{url}            % simple URL typesetting
\usepackage{booktabs}       % professional-quality tables
\usepackage{amsfonts}       % blackboard math symbols
\usepackage{nicefrac}       % compact symbols for 1/2, etc.
\usepackage{microtype}      % microtypography
\usepackage{graphicx}
\usepackage{float}
\usepackage{framed}
\usepackage{mathrsfs}
\usepackage{amsmath}
\usepackage{xcolor}
\usepackage{algorithm}
\usepackage{algorithmic}
\usepackage{bm}
\usepackage{subfigure}
\usepackage{epsfig}
\usepackage{epstopdf}
\usepackage{amsthm}
\usepackage{cite}
\usepackage{amssymb}
\usepackage{textcomp}

\title{QMNet: Importance-Aware Message Exchange for Decentralized Multi-Agent Reinforcement Learning}

\author{
    Xiufeng~Huang,~\IEEEmembership{Student~Member,~IEEE,}
    Sheng~Zhou,~\IEEEmembership{Member,~IEEE}
 \IEEEcompsocitemizethanks{\IEEEcompsocthanksitem The work is sponsored in part by the Natural Science Foundation of China (No. 62341108, No. 62022049, No. 62111530197) and Hitachi Ltd. Part of the paper has been accepted by GLOBECOM 2022 \cite{gc2022marl}.

 Xiufeng Huang and Sheng Zhou are with Beijing National Research Center for Information Science and Technology, Department of Electronic Engineering, Tsinghua University, Beijing, China.
E-mail: huangxf18@mails.tsinghua.edu.cn, sheng.zhou@tsinghua.edu.cn}% <-this % stops an unwanted space
}
\begin{document}

\IEEEtitleabstractindextext{%

\begin{abstract}

To improve the performance of multi-agent reinforcement learning under the constraint of wireless resources, we propose a message importance metric and design an importance-aware scheduling policy to effectively exchange messages. The key insight is spending the precious communication resources on important messages. The message importance depends not only on the messages themselves, but also on the needs of agents who receive them. Accordingly, we propose a query-message-based architecture, called QMNet. Agents generate queries and messages with the environment observation. Sharing queries can help calculate message importance. Exchanging messages can help agents cooperate better. Besides, we exploit the message importance to deal with random access collisions in decentralized systems. Furthermore, a message prediction mechanism is proposed to compensate for messages that are not transmitted. Finally, we evaluate the proposed schemes in a traffic junction environment, where only a fraction of agents can send messages due to limited wireless resources. Results show that QMNet can extract valuable information to guarantee the system performance even when only $30\%$ of agents can share messages. By exploiting message prediction, the system can further save $40\%$ of wireless resources. The importance-aware decentralized multi-access mechanism can effectively avoid collisions, achieving almost the same performance as centralized scheduling.

\end{abstract}

}

\maketitle
% keywords can be removed
%\begin{IEEEkeywords}
%Edge computing, Machine learning, Data filtering
%\end{IEEEkeywords}

\IEEEraisesectionheading{\section{Introduction}\label{sec:introduction}}

\IEEEPARstart{R}{ecently}, reinforcement learning has made great progress combined with deep learning and provided many intelligent services, such as vehicular network \cite{althamary2019survey}, robotic control \cite{lillicrap2015continuous} and game playing \cite{mnih2015human}. Although reinforcement learning has achieved great success \cite{silver2016mastering}, to enable more intelligent services in real-world environments, one key is the cooperation among intelligent agents and thus many researches have focused on multi-agent reinforcement learning (MARL), including MADDPG \cite{lowe2017multi}, COMA \cite{COMA}, VDN \cite{sunehag2018vdn}, QMIX \cite{rashid2018qmix}, MAAC \cite{iqbal2019actor} and QTRAN \cite{QTRAN}. Compared with single-agent reinforcement learning, MARL is faced with partial observability and the loss of stationarity of the environment \cite{hernandez2017survey}\cite{laurent2011world}, and thus independent processing of every agent without cooperation can hardly perform well \cite{tan1993multi}\cite{lanctot2017unified}. Therefore, communication becomes the key part of multi-agent systems to solve these challenges and improve system performance \cite{jim2001communication}. With communication, agents can share the information of their observations and decisions, which is encoded in the messages generated by neural networks, such as RIAL and DIAL \cite{foerster2016learning}, CommNet \cite{commnet} and BiCNet \cite{peng2017multiagent}. However, with the explosive development of intelligent devices \cite{7498684}, the cooperation in large-scale future intelligent systems and other emerging intelligent services will require delivering a huge amount of data wirelessly. Therefore, wireless resources will inevitably become scarce and thus communication is no doubt the bottleneck of large-scale MARL systems. On the other hand, not every agent can provide important information for the cooperation of MARL systems. Exchanging all messages may cause unnecessary waste of communication resources with marginal performance improvement.

This paper targets at improving the performance of MARL systems with limited communication resources. Under the constraint of bandwidth, only part of agents can send their messages to other agents. Therefore, the first problem is how to schedule agents to exchange their messages with each other. The idea is to select the most important messages to share and thus the problem turns to how to evaluate the importance of messages. Recent researches have proposed some methods to define the message importance. In individually inferred communication (I2C) \cite{NEURIPS2020_fb2fcd53}, the message importance is the difference between value functions with and without the receiving message as model input. Some researches exploit neural-network-based attention model to generate the message importance, including emergent communication (EC) \cite{emergent2020}, attentional communication (ATOC) \cite{jiang2018learning}, intention sharing (IS) \cite{kim2021communication}, vertex attention interaction network (VAIN) \cite{hoshen2017vain}, targeted multi-agent communication (TarMAC) \cite{das2019tarmac}, intrinsic motivated multi-agent communication (IMMAC) \cite{sun2021intrinsic}. IC3Net \cite{singh2018learning} uses a neural-network-based gate mechanism to decide when to communicate for both cooperative and competitive tasks, which can also be regarded as the message importance. HAMMER \cite{gupta2021hammer} considers a powerful central agent to learn what messages should be sent. FlowComm \cite{du2021learning} learns communication strategy according to the network topology. MAGIC \cite{niu2021multi} and  GAXNet \cite{yun2021attention} learn graph-based attention to measure message importance. In \cite{jiang2019graph}\cite{su2020counterfactual}, graph-convolutional is applied to learn the relationships among the agents and generate the message importance. 

To design the importance-aware communication scheme, the first problem is how to define the message importance. From our perspective, the message importance depends not only on the messages themselves, but also on the needs of other agents who receive them. Therefore, we propose QMNet (query-message network), designing a query mechanism to enable agents to express what they can provide and what they need before sending messages. Different from recent works, QMNet costs a few communication resources to improve the estimation of message importance. Queries can be regarded as compressed messages, encoding part of information, with which agents can calculate the importance of messages with small communication overhead. With the message importance evaluated by the query mechanism, the agents can select important messages for transmission to save communication resources and perform weighted averaging for message aggregation. Besides, the information encoded in queries can be utilized to help generate messages, which enables agents to send shorter messages while guaranteeing the system performance. In our work, we define the message importance as their influence on the MARL model output, i.e., more change of the model output (such as Q-value, action probability) after receiving updated messages means larger importance of these messages. 

In practical scenarios, the MARL systems may be decentralized. Without the centralized scheduler, agents will be faced with collisions while accessing the same wireless channel. The access collisions will waste communication resources and the problem is how to avoid these collisions in the decentralized MARL system. Some researches \cite{zhang2018fully}\cite{qu2020scalable} focus on designing networked MARL systems to share information among decentralized agents but overlooked the communication collision problem. For general wireless communication systems, there are some decentralized multi-access mechanisms, such as carrier sense multiple access (CSMA) and semi-persistent scheduling (SPS). To take advantage of the message-importance-based scheduling policy mentioned above, SchedNet \cite{kim2019learning} provides an insight to integrate message-importance-based scheduling with CSMA, by using message importance to generate the backoff time. The agent with larger message importance will use shorter backoff time and thus it can have higher success probability of accessing the wireless channel. In this paper, we utilize the query mechanism to propose an importance-aware decentralized multi-access mechanism to avoid collisions. The queries are periodically sent for agents to calculate message importance. The broadcasted queries provide common knowledge, used for coordination, among agents. With the common knowledge, i.e., the calculated message importance, agents can agree on the same scheduling policy and send their messages in proper order. The proposed importance-aware decentralized multi-access mechanism can successfully avoid collisions while sending messages.

Although the above agent scheduling and decentralized multi-access mechanism can select important messages to transmit, there is still information loss because some agents are unable to update their messages due to limited wireless resources. The loss of information may result in severe performance degradation. Therefore, we introduce message prediction into the MARL system to compensate for the lost information. It is inspired by the fact that the agents with less message importance have higher probability to be stable in terms of states and decisions. Therefore we can predict the messages of these agents according to their history messages and the knowledge of environments that is encoded in the reinforcement learning model. When not receiving new messages, the agents will use the predicted messages as the input of the reinforcement model. The predicted messages can significantly reduce the information loss resulted by limited communication resources. Finally, with the message prediction block, the performance of the MARL system becomes robust with respect to the communication resources.

To summarize, in this paper we design importance-aware message exchange and message prediction schemes to optimize the decentralized MARL systems with limited communication resources. In particular, our contributions include: 

 \begin{itemize}
    \item Design an agent scheduling policy and decentralized multi-access mechanism with the query mechanism. The queries are periodically broadcasted in the wireless channel to help select important messages for exchange and perform weighted averaging for message aggregation. With the scheduling policy generated by the query mechanism, agents can send messages in order, avoiding collisions during medium access. To reduce extra communication overhead resulted by sending queries, we further propose a query-based message generation block to better utilize the information encoded in queries and enable agents to send shorter messages.
    \item Propose a definition of message importance as the influence on the model output and derive a simplified formulation of message importance. The simplified message importance is easier to calculate and does not need online training, which makes it suitable for deployment in dynamic environments. Simulation results show that the messages selected by message importance can significantly improve the system performance under the constraint of wireless resources.
    \item Design message prediction block for the agents that are not scheduled to share messages. In fact, the agents not scheduled have high probability to be stable, and their messages do not have significant change. Therefore, by exploiting history information and the knowledge of the environment learned by the reinforcement learning model, these messages can be predicted with high accuracy and the information loss due to limited wireless resources can be reduced.
    
 \end{itemize}

The rest of this paper is organized as follows. In Section \ref{sec:sysmod} we introduce the system model of the considered MARL system, including the reinforcement learning model and the setting of message exchange. In Section \ref{sec:query} we introduce the proposed query mechanism, including query-based message generation and aggregation. In Section \ref{sec:importance} we propose the definition of message importance and the derivation of its simplified formulation, with which we design a scheduling policy for centralized MARL systems and a multi-access mechanism for decentralized MARL systems. Section \ref{sec:predict} describes the algorithm of the message prediction block. Experiments results are provided in Section \ref{sec:exp}. The paper is concluded in Section \ref{sec:conclusion}.

\section{System Model}\label{sec:sysmod}

%\subsection{System model}

\begin{figure}[t]
%\vspace{-1.5em}
\begin{center}
\centerline{\includegraphics[width=0.95\columnwidth]{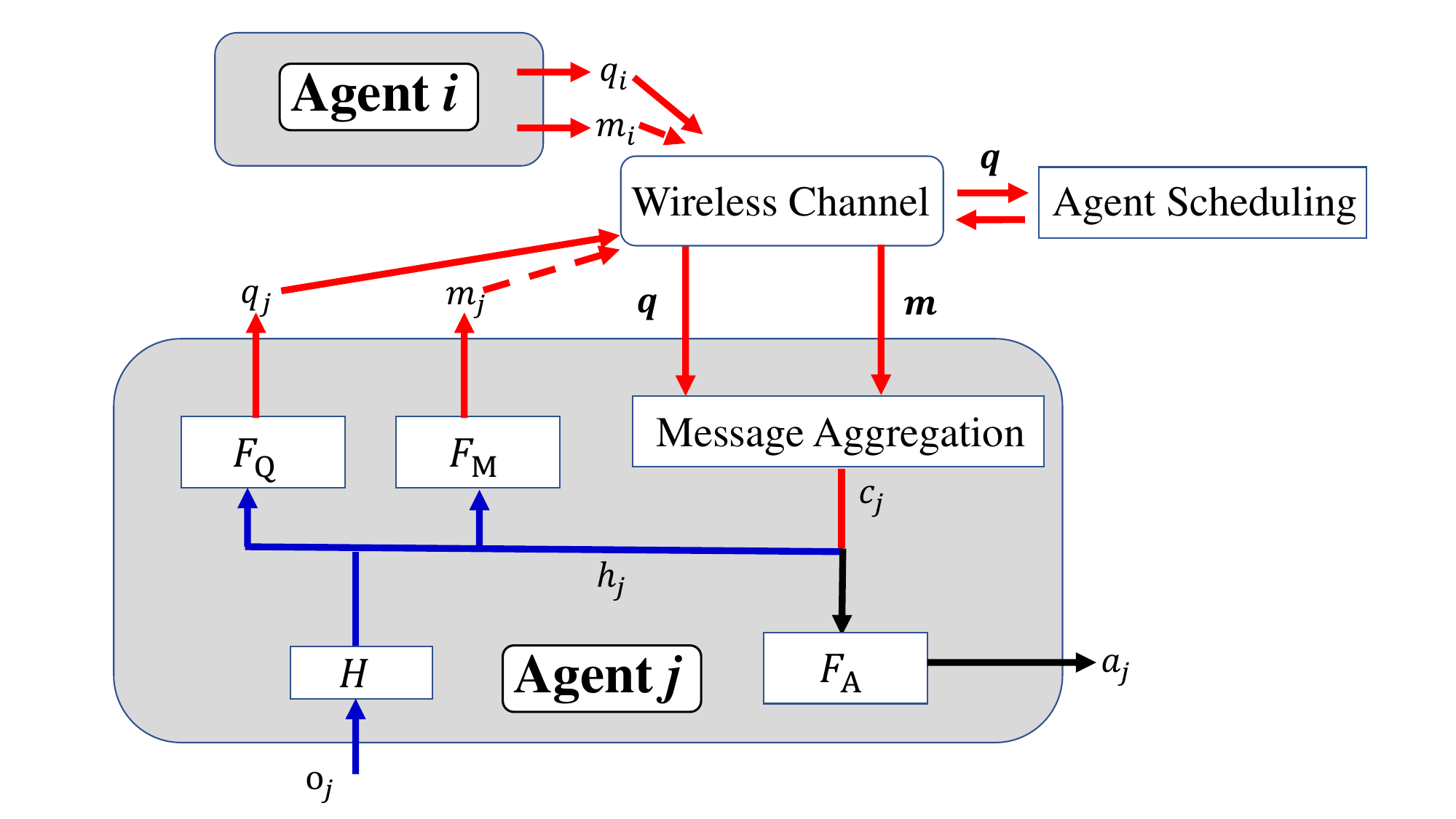}}
\caption{The MARL model adopted in this paper.}
\label{fig:model}
\end{center}
\vspace{-1em}
\end{figure}

The reinforcement learning model designed in this work, called QMNet, is shown in Fig. \ref{fig:model}, similar to \cite{commnet}, except that we propose to add a message block $F_{\rm M}$ and a query block $F_{\rm Q}$, which will be introduced in Sec. \ref{sec:query}. Take the $j$-th agent for example in Fig. \ref{fig:model}, in every round, agent $j$ obtains its observation $\boldsymbol{o}_j$ and generates its hidden state $\boldsymbol{h}_j = H(\boldsymbol{o}_j)$ with a neural network $H$. Next, agent $j$ generates message $\boldsymbol{m}_j = F_{\rm M}(\boldsymbol{h}_j)$ and query $\boldsymbol{q}_j = F_{\rm Q}(\boldsymbol{h}_j)$, which are real-valued vectors encoding the information needed by other agents for cooperation. All agents share their queries in the query phase. Upon receiving the queries, agents or centralized controllers calculate the message importance of every agent, i.e., how these messages can impact other agents. In the message phase, with the message importance, agents send their messages according to the centralized scheduling policy or importance-aware decentralized multi-access mechanism. After receiving updated messages, agent $j$ aggregates these messages by weighted averaging to obtain $\boldsymbol{c}_j$. The details of the message aggregation will be introduced in Sec. \ref{sec:query}. Finally, agent $j$ generates its action $a_j = F_{\rm A}(\boldsymbol{h}_j,\boldsymbol{c}_j)$ with the hidden unit $\boldsymbol{h}_j$ and aggregated message $\boldsymbol{c}_j$ via the action block $F_{\rm A}$. The reinforcement learning model of every agent, including hidden layer $H$, query block $F_{\rm Q}$, message block $F_{\rm M}$ and action block $F_{\rm A}$, which are all fully connected layers. The convergence of the MARL model depends on the training policy and we exploit asynchronous advantage actor-critic (A3C) \cite{A3C} to train it. The system is implemented by centralized training and decentralized execution (CTDE). The parameters of the learning model are shared among all agents and the training maximizes individual reward like IC3Net \cite{singh2018learning}, instead of global reward.

In the MARL system under consideration, there are $N$ agents coordinated to perform a reinforcement learning task, which is described as a tuple $\langle N, \boldsymbol{S}, \boldsymbol{A}, \boldsymbol{r}, P, \boldsymbol{O}, \boldsymbol{M}, \boldsymbol{Q}, T, N{\rm c} \rangle$. The element symbols are explained as follows. 

$\boldsymbol{S}$ is the state space of the whole environment. $\boldsymbol{A}$ is the action space of all agents. $\boldsymbol{r} = \{r_1,r_2,\cdots,r_N\}$ is the set of rewards of the reinforcement learning task. $P : \boldsymbol{S} \times \boldsymbol{A} \rightarrow \boldsymbol{S}$ is the state transition function of the environment. In the considered multi-agent system, agents only have partial observations of the environment instead of global view. $\boldsymbol{O} = \{\boldsymbol{O}_i\}_{i=1, \dots, N}$ is the space of all agents' observations. For coordination, the agents generate messages according to their observations with the learning model and share the messages through wireless channels. $\boldsymbol{M} = \{\boldsymbol{M}_i\}_{i=1, \dots, N}$ is the space of messages. In our work, we propose QMNet (Fig. \ref{fig:model}), exploiting a query mechanism to optimize the communication scheme of the MARL system, which will be introduced in Sec. \ref{sec:query} in detail. Agents will broadcast queries before sending messages and $\boldsymbol{Q} = \{\boldsymbol{Q}_i\}_{i=1, \dots, N}$ is the space of queries. The numbers of elements of message and query are $s_{\rm M}$ and $s_{\rm Q}$ (each element has the size of 4 bytes). 

\begin{figure}[t]
%\vspace{-1.5em}
\begin{center}
\centerline{\includegraphics[width=0.9\columnwidth]{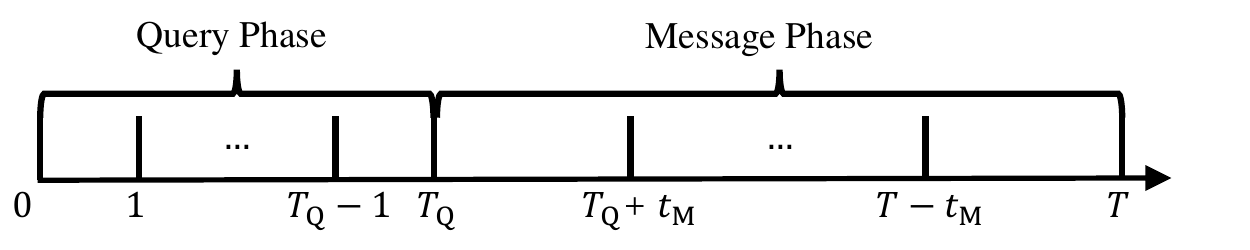}}
\caption{Query phase and message phase in one round.}
\label{fig:time}
\end{center}
\vspace{-1em}
\end{figure}

The system is time-slotted and every agent takes one action in every round, including $T$ time slots, as shown in Fig. \ref{fig:time}. For communication, a round is divided into two phases, \textbf{the query phase} and \textbf{the message phase}. In the query phase, which includes $T_{\rm Q}$ time slots, the agents share their queries and every query needs $t_{\rm Q}=1$ time slot for transmission. Because the query has a much smaller size, i.e., fewer elements in the vector, than the message, and only costs one time slot for transmission, we assume that $T_{\rm Q}$ time slots in the query phase are sufficient for the transmission of all agents' queries. If there are more than $T_{\rm Q}$ agents in the MARL system, the excessive agents will not obtain the opportunity to communication and thus can be regarded as part of the environment, instead of the agents of the MARL system. Agents can periodically take fixed time slots to broadcast queries in every round. Compared with query-free communication schemes, QMNet requires extra communication resources to send queries. Sec. \ref{sec:query} will discuss this overhead and propose ways to reduce it while guaranteeing the system performance.

Next, agents share messages in the message phase, with $T_{\rm M} = T - T_{\rm Q}$ time slots. Every message needs $t_{\rm M}$ time slots for transmission. Due to the limited wireless resources, the time slots in the message phase are only enough for the transmission of messages of $N_{\rm c} = \lfloor \frac{T_{\rm M}}{t_{\rm M}} \rfloor$ agents (if the system is decentralized and collisions happen during wireless channel access, the number of sent messages will be smaller). We propose a scheduling policy to select agents to take the wireless channel for centralized MARL systems and an importance-aware channel access mechanism for decentralized MARL systems in Sec. \ref{sec:importance}.

%Add description of multi-access

The objective of the MARL system is maximizing the summation of rewards with given communication resources:
\begin{align}
&&\max\limits_{\boldsymbol{\mathcal{M}}}\ &\sum\limits_{i=1}^{N} r_i \label{eq:obj1}\\
&&\text{s.t.}\ &\boldsymbol{\mathcal{M}} \subseteq \{\boldsymbol{m}_1, \boldsymbol{m}_2, \cdots, \boldsymbol{m}_N\} \nonumber\\
&&\ &|\boldsymbol{\mathcal{M}}|t_M \le T_M \nonumber\\
%\end{array}
\nonumber
\end{align}
$\boldsymbol{\mathcal{M}}$ is the set of selected messages for sharing. However, the relationship between reward $\boldsymbol{r}$ and shared messages $\boldsymbol{\mathcal{M}}$ is difficult to be explicitly expressed. Therefore, we hope to find the importance measure for messages, $I(\boldsymbol{m})$, which can represent the improvement of reward $\sum\limits_{i=1}^{N} r_i$ brought by sharing message $\boldsymbol{m}$. In this way, the optimization problem (\ref{eq:obj1}) can be turned to (\ref{eq:obj2}) as follows,
\begin{align}
&&\max\limits_{\boldsymbol{\mathcal{M}}}\ &\sum\limits_{\boldsymbol{m} \in \boldsymbol{\mathcal{M}}} I(\boldsymbol{m}) \label{eq:obj2}\\
&&\text{s.t.}\ &\boldsymbol{\mathcal{M}} \subseteq \{\boldsymbol{m}_1, \boldsymbol{m}_2, \cdots, \boldsymbol{m}_N\} \nonumber\\
&&\ &|\boldsymbol{\mathcal{M}}|t_M \le T_M \nonumber\\
%\end{array}
\nonumber
\end{align}
The calculation of $I(\boldsymbol{m})$ will be introduced in Sec. \ref{sec:importance}. The solution of (\ref{eq:obj2}) is obvious and it only needs to select $\lfloor \frac{T_{\rm M}}{t_{\rm M}} \rfloor$ messages with the highest message importance for sharing.

Tab. \ref{tab:notation} summarizes the main notations in the paper.
\begin{table}
  
  \begin{center}
    \caption{Summary of main notations}
    \begin{tabular}{cc} % <-- Alignments: 1st column left, 2nd middle and 3rd right, with vertical lines in between
      Notation & Explanation\\ \hline
      $\boldsymbol{o}$ & Agent's observation of environment \\
      $\boldsymbol{h}$ & Hidden state generated by RL model \\
      $\boldsymbol{m}$ & Message generated by agent for sharing \\
      $\boldsymbol{m}'$ & Message for MARL model input when not\\
      					& receiving updated message \\
      $\boldsymbol{q}$ & Query generated by agent for sharing \\
      $\boldsymbol{c}$ & Integrated message \\
      $\boldsymbol{\theta}$ & Parameters of RL model \\
      $I_j(\boldsymbol{m})$ & Message importance for agent $j$ \\
      $I(\boldsymbol{m})$ & Message importance for the MARL system \\
      $s_{\rm M},s_{\rm Q}$ & Size of message and query\\ 
      $t_{\rm M},t_{\rm Q}$ & Number of time slots used for\\
      						& sending message and query\\ \hline
    \end{tabular}
  \end{center}
  \label{tab:notation}
\end{table}

\section{Query Mechanism}\label{sec:query}

\subsection{Message Block and Query Block}

In recent work, such as \cite{commnet}, the MARL system uses hidden state of deep reinforcement learning model as shared messages. In our work, we propose QMNet, adding a message block and a query block, which are fully connected layers, to improve the performance of the MARL system as shown in Fig. \ref{fig:model}. 

The hidden state $\boldsymbol{h}_j$ does not only encode the information needed for other agents, but also encodes the information needed for making the $j$-th agent's own decisions. Therefore, sending hidden units will cost excessive communication resources. In QMNet, we add the message block $F_{\rm M}$, a neural network, to extract the information needed for sharing among agents and save communication resources. For example, in an intersection scheduling task, in which vehicles pass through an intersection and need to avoid collisions, the hidden state and the message can encode the information about the vehicle's location, planned route ,and speed. If the vehicle is far from the intersection, then the message may be unnecessary to encode some information, such as its planned route.

The query is also generated by the hidden unit and it can be regarded as the compressed message. For example, in the intersection scheduling task, the message encodes the information of a specific location but the query only needs to encode whether the vehicle is close to the intersection. And the queries can encode the information about what the agents can provide and what the agents need. We propose the query block $F_{\rm Q}$ to cost fewer communication resources (compared with messages) to share information among agents. With the information encoded in queries, agents can realize the message importance of other agents before sending messages and make better decisions of communication scheduling, which will be introduced in detail in Sec. \ref{sec:importance}.

\subsection{Query-Based Message Aggregation}

Next, we focus on the aggregation of messages. In some recent works, the aggregated message $\boldsymbol{c}_j$ used as the model input is average over all agents' messages, i.e.,
\begin{equation}\label{eq:sum}
\boldsymbol{c}_j = \frac{1}{N-1} \sum\limits_{i \neq j} \boldsymbol{m}_i, 
\end{equation}
where the messages from different agents have the same weights. It means that these messages have the same contribution to agent $j$'s decision making.

However, in practical scenarios, these messages may not have the same contribution. The importance of received messages from agents can be different, depending on their positions, targets or other information that is significant for the reward of the MARL system. Accordingly, we exploit the query mechanism to help with the aggregation of messages. A query represents the needs of the agent who sends it out. We propose a new message aggregation method along with the query mechanism and $\boldsymbol{c}_j$ is defined as
\begin{equation}\label{eq:aggre}
\boldsymbol{c}_j = \frac{1}{N-1} \sum\limits_{i \neq j} \boldsymbol{q}^\mathrm{T}_j \boldsymbol{q}_i \boldsymbol{m}_i,
\end{equation}
where $\boldsymbol{q}_i$ is the query vector generated by the $i$-th agent. After enough training of the RL model, the larger dot product of the queries from two agents can represent strong mutual importance, and thus larger weights are adopted in their message aggregation processes (\ref{eq:aggre}). In this way, we propose a weighted averaging method for message aggregation with queries.

%Add description of message aggregation

%In the proposed MARL system, the designed query has much smaller size than the message, which ensures the system benefits from broadcasting queries of all agents without too much extra communication overhead. 
%Besides providing weights for message aggregation, broadcasting queries before message sharing can also help schedule agents to transmit their messages according to the information encoded in the queries. The details of how to select important messages are described in the next section. 

\subsection{Query-Based Message Generation}

Although queries help schedule agents and enable weighted message aggregation, they result in extra communication overhead. Without queries, every agent only needs to send a message vector with $s_{\rm M}$ bytes. After introducing queries, every agent needs to send $s_{\rm M} + s_{\rm Q}$ bytes in every round. Can we reduce this communication overhead while guaranteeing the performance of the message-based MARL model? The answer is yes.

Notice that the query $\boldsymbol{q}=F_{\rm Q}(\boldsymbol{h})$ and the message $\boldsymbol{m}=F_{\rm M}(\boldsymbol{h})$ both encode the agent's information needed by other agents. The query $\boldsymbol{q}$ may include part of information of the message $\boldsymbol{m}$ because they are both generated with $\boldsymbol{h}$. Therefore, with the information encoded in $\boldsymbol{q}$, the agent can send shorter message $\boldsymbol{m}'$ while achieving almost the same performance as sending $\boldsymbol{m}$, by exploiting $\boldsymbol{q}$ and $\boldsymbol{m}'$ to generate $\boldsymbol{m}$ at the receiver side. Based on this insight, we propose a query-based message generation function $G$. The agent can generate and send message $\boldsymbol{m}'=F_{\rm M'}(\boldsymbol{h})$ with small size $s_{\rm M'}$ (e.g., $s_{\rm M'} = s_{\rm M} - s_{\rm Q}$). At the receiver side, the message used for message aggregation is $\boldsymbol{m}=G(\boldsymbol{q},\boldsymbol{m}')$, which is reconstructed with the information from $\boldsymbol{q}$ and $\boldsymbol{m}'$. 

Combined with the query-based message generation block, the complete message aggregation becomes 
\begin{equation}\label{eq:aggre}
\boldsymbol{c}_j = \frac{1}{N-1} \sum\limits_{i \neq j} \boldsymbol{q}^\mathrm{T}_j \boldsymbol{q}_i G(\boldsymbol{q}_i,\boldsymbol{m}_i).
\end{equation}
In this way, the query becomes part of the message and the agent can send shorter messages to avoid the overhead resulted by introducing the query mechanism. Simulation results in Sec. \ref{sec:exp} show there is no significant performance degradation resulted by sending shorter messages. 

\section{Agent Scheduling Policy and Decentralized Multi-Access Mechanism}\label{sec:importance}

\subsection{Message Importance Measure}
One of the key issues of the MARL system is the communication bottleneck. In practice, there can be a large number of agents broadcasting their messages in a wireless channel, which may not be feasible in time due to the limited wireless resources. Therefore, it is vital to reduce the communication cost of the system. In our work, we schedule a subset of agents to broadcast their messages in every round. Specifically, in one round, only $N_{\rm c}$ agents can be allocated with time slots to broadcast their message over the wireless channel. The problem is which $N_{\rm c}$ out of all agents should be selected. To solve this problem, we first need to define the importance of message.

If properly trained, the change of the reinforcement learning model output resulted by the shared messages can improve the performance of the whole system. Therefore, the messages affecting more on the model output should have larger importance. We first consider the message importance for one agent. For example, upon receiving message $\boldsymbol{m}_i$, the model output of the $j$-th agent is $f(\boldsymbol{m}_i;\boldsymbol{\theta}_j, \boldsymbol{o}_j, \boldsymbol{m}^{\rm c})$, where $\boldsymbol{\theta}_j$ is its model parameters, $\boldsymbol{o}_j$ is its observation and $\boldsymbol{m}^{\rm c}$ includes the messages receiving from other agents. If agent $j$ does not receive message $\boldsymbol{m}_i$, the message used for model input can be said $\boldsymbol{m}'_i$ (e.g. the history message from agent $i$) and the model output is $f(\boldsymbol{m}'_i;\boldsymbol{\theta}_j, \boldsymbol{o}_j, \boldsymbol{m}^{\rm c})$. Then we can obtain the message importance of $\boldsymbol{m}_i$ for agent $j$ as
\begin{equation}\label{eq:Ij} 
I_j(\boldsymbol{m}_i) = \left\| f(\boldsymbol{m}_i;\boldsymbol{\theta}_j, \boldsymbol{o}_j, \boldsymbol{m}^{\rm c}) - f(\boldsymbol{m}'_i;\boldsymbol{\theta}_j, \boldsymbol{o}_j, \boldsymbol{m}^{\rm c}) \right\|.
\end{equation}
In a MARL system that continuously exchanges messages, the message distance, i.e., $\left\| \boldsymbol{m}_i - \boldsymbol{m}'_i \right\|$, is small. Therefore the message importance can be approximated by 

\begin{align}
I_j(\boldsymbol{m}_i) &\approx \left\| \frac{\partial f}{\partial \boldsymbol{c}}\Big|_{\boldsymbol{c}=\boldsymbol{c}_j}  \frac{\partial \boldsymbol{c}}{\partial \boldsymbol{m}}\Big|_{\boldsymbol{m}=\boldsymbol{m}_i} (\boldsymbol{m}_i-\boldsymbol{m}'_i) \right\| \\
&=\left\| \frac{\partial f}{\partial \boldsymbol{c}}\Big|_{\boldsymbol{c}=\boldsymbol{c}_j} \boldsymbol{q}^\mathrm{T}_j \boldsymbol{q}_i(\boldsymbol{m}_i-\boldsymbol{m}'_i) \right\|,\label{eq:Ij2} 
\end{align}
where the calculation of $I_j(\boldsymbol{m}_i)$ needs the information from agent $j$, i.e., $\frac{\partial f}{\partial \boldsymbol{c}}|_{\boldsymbol{c}=\boldsymbol{c}_j}$, which requires extra communication resources for sharing among agents. To estimate $I_j(\boldsymbol{m}_i)$ without sharing gradient $\frac{\partial f}{\partial \boldsymbol{c}}|_{\boldsymbol{c}=\boldsymbol{c}_j}$, we treat it as a random vector, assuming the uniform distribution of the direction and identical distribution of the magnitude, i.e., 
\begin{align}\label{eq:idl} 
&\mathbb{P}\left(\left\| \frac{\partial f}{\partial \boldsymbol{c}}\Big|_{\boldsymbol{c}=\boldsymbol{c}_a} \right\| = g;\boldsymbol{c}_a\right) \nonumber\\
= &\mathbb{P}\left(\left\| \frac{\partial f}{\partial \boldsymbol{c}}\Big|_{\boldsymbol{c}=\boldsymbol{c}_b} \right\| = g;\boldsymbol{c}_b\right), \forall g \in \mathbb{R}^{*}, \boldsymbol{c}_a,\boldsymbol{c}_b \in \textbf{M}.
\end{align}
Then, we obtain the message importance by taking the expectation
\begin{align}
I_j(\boldsymbol{m}_i) &\approx \mathbb{E}_{\frac{\partial f}{\partial \boldsymbol{c}}\big|_{\boldsymbol{c}=\boldsymbol{c}_j}} \left[\left\| \frac{\partial f}{\partial \boldsymbol{c}}\Big|_{\boldsymbol{c}=\boldsymbol{c}_j} \boldsymbol{q}^\mathrm{T}_j \boldsymbol{q}_i (\boldsymbol{m}_i-\boldsymbol{m}'_i) \right\|\right] \\
&=\frac{\int^{2\pi}_{0} |{\rm cos}\theta| {\rm d}\theta}{2\pi} \int^{\infty}_{0} gp(g)\|\boldsymbol{q}^\mathrm{T}_j \boldsymbol{q}_i (\boldsymbol{m}_i-\boldsymbol{m}'_i)\|{\rm d}g \\
&= C\|\boldsymbol{q}^\mathrm{T}_j \boldsymbol{q}_i (\boldsymbol{m}_i-\boldsymbol{m}'_i)\|,\\
\nonumber
\end{align}
where $p(g) = \mathbb{P}\left(\left\| \frac{\partial f}{\partial \boldsymbol{c}}\Big|_{\boldsymbol{c}=\boldsymbol{c}_j} \right\| = g;\boldsymbol{c}_j\right)$ and $C = \frac{\int^{2\pi}_{0} |{\rm cos}\theta| {\rm d}\theta}{2\pi} \int^{\infty}_{0} gp(g){\rm d}g$. Note that $C$ is a constant independent of $j$ and $i$. Therefore, to compare the importance of messages, we can simplify $I_j(\boldsymbol{m}_i)$ into
\begin{equation}\label{eq:Ij3} 
I_j(\boldsymbol{m}_i) = \|\boldsymbol{q}^\mathrm{T}_j \boldsymbol{q}_i(\boldsymbol{m}_i-\boldsymbol{m}'_i)\|.
\end{equation}
Eq. (\ref{eq:Ij3}) shows an approximate method for calculating the message importance without extra information sharing by other agents. The key idea is that the importance depends on the change of messages as well as the weights of message aggregation.

In fact, after simplification of the message importance, the message $m$ with higher $I_j(m)$ shown in (\ref{eq:Ij3}) is not always more important than other messages. However, a set of messages with higher message importances formulated by (\ref{eq:Ij3}) can be more important than other sets of messages because (\ref{eq:Ij3}) is simplified by expectation. Therefore, selecting a set of important messages defined by (\ref{eq:Ij3}) can still help improve the performance of MARL systems and the simulation results to support it will be shown in Sec. \ref{sec:exp}.

Note that the message importance in (\ref{eq:Ij3}) depends on how we choose the alternative input $\boldsymbol{m}'_i$. For example, if we exploit zero vector $\boldsymbol{m}'_i = \boldsymbol{0}$, the message importance will be 
\begin{equation}\label{eq:influ} 
I_j(\boldsymbol{m}_i) = \|\boldsymbol{q}^\mathrm{T}_j \boldsymbol{q}_i \boldsymbol{m}_i\|,
\end{equation}
namely the influence-based message importance. Another choice is using history message $\boldsymbol{m}^{\mathrm{o}}_i$ (of the last round when agent $i$ is scheduled) as model input $\boldsymbol{m}'_i$, which obtains the message importance
\begin{equation}\label{eq:diff} 
I_j(\boldsymbol{m}_i) = \|\boldsymbol{q}^\mathrm{T}_j \boldsymbol{q}_i(\boldsymbol{m}_i-\boldsymbol{m}^{\mathrm{o}}_i)\|,
\end{equation}
namely the difference-based message importance. In Sec. \ref{sec:predict} we will use predicted message to replace $\boldsymbol{m}'_i$ to achieve better performance.

%Centralized Scheduling, Decentralized: query phase, random access, importance-aware access

\subsection{Centralized Scheduling}

\begin{figure}[t]
%\vspace{-1.5em}
\begin{center}
\centerline{\includegraphics[width=0.95\columnwidth]{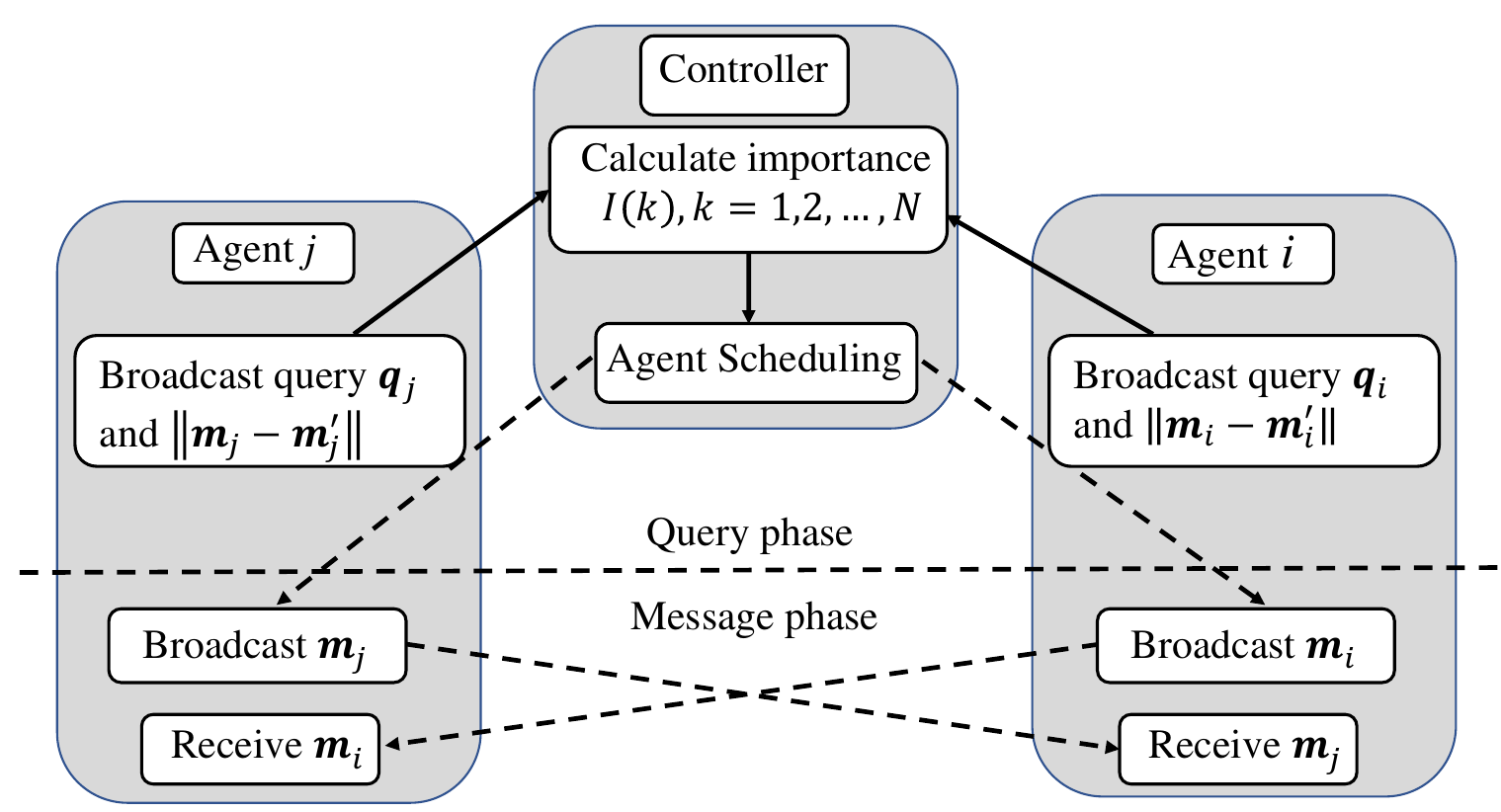}}
\caption{The workflow of centralized scheduling.}
\label{fig:flow}
\end{center}
%\vspace{-3em}
\end{figure}

After proposing the message importance measure, the next step is scheduling agents to send important messages, in order to achieve better MARL system performance under the constraint of wireless resources. Fig. \ref{fig:flow} shows the workflow of agent scheduling for message exchange. After generating messages and queries, a centralized controller will schedule the agents to send their queries in order in the query phase. Eq. (\ref{eq:Ij3}) shows that the calculation of message importance needs not only the queries, but also the message distance $\|\boldsymbol{m}_i-\boldsymbol{m}'_i\|, i=1,2,\cdots,N$. Therefore, in the query phase, the message distance, only a floating point number, is integrated into the packet of query for broadcasting. The message importance of an agent is the summation of its impact on all other agents. Taking agent $i$ as example, its message importance is
\begin{equation}\label{eq:Ij} 
I(i) = \frac{1}{N-1} \sum\limits_{j \neq i} I_j(\boldsymbol{m}_i).
\end{equation}

After receiving the queries, the centralized controller calculates the message importance of all agents and broadcasts the scheduling policy. The $N_{\rm c}$ agents with the highest message importance will be scheduled to take the wireless channel in the order of message importance during the message phase.

\subsection{Importance-Aware Decentralized Multi-Access}

In some scenarios, such as formation of UAVs or autonomous vehicles, the communication system is implemented with ad hoc network and thus no centralized controller can be used to schedule the communication. In this subsection, we will consider the communication mechanisms for these decentralized cases. In these decentralized MARL systems, the agents can use some random access mechanisms, such as CSMA, to contend for wireless resources. However, the contention will result in the waste of wireless resources. With more agents in the MARL system, the system will have higher degree of decentralization and channel access will be more difficult. To take the advantage of the proposed message importance, we propose an importance-aware decentralized multi-access mechanism to avoid access collision. The mechanism includes two phases: query phase and message phase, described as follows.

\textbf{Query phase}: As introduced in Sec. \ref{sec:sysmod}, there are $T_{\rm Q}$ time slots for broadcasting queries and it is assumed that $T_{\rm Q}$ time slots are enough for the transmission of all agents' queries. The time slots in the query phase are indexed in order (from $1$ to $T_{\rm Q}$). Because the transmission of query is periodic, a specific agent can take a fixed time slot to send its query in the query phase of every round. 

The problem is how these agents take their time slots without collisions. When an agent enters the MARL system, it can sense the wireless channel and find the idle time slots of the query phase in the current round. Then this agent can randomly take one idle time slot and send its query in this time slot in the following rounds. 

The next problem is that collisions may happen when more than one agent enters the MARL system in the same round and they take the same idle time slot to send queries. If the communication of the system is full duplex, these agents can sense the collisions of sending queries. The agents that encounter collisions can continue to randomly select idle time slots in the following rounds until no collision happens. If an agent sends a query successfully (i.e., without collision) in a time slot, it will always take this time slot to send its query in the following rounds.  

\begin{figure} [t]
	\centering
	\subfigure[Acks flow when collision happens.]{
		\includegraphics[scale=0.4]{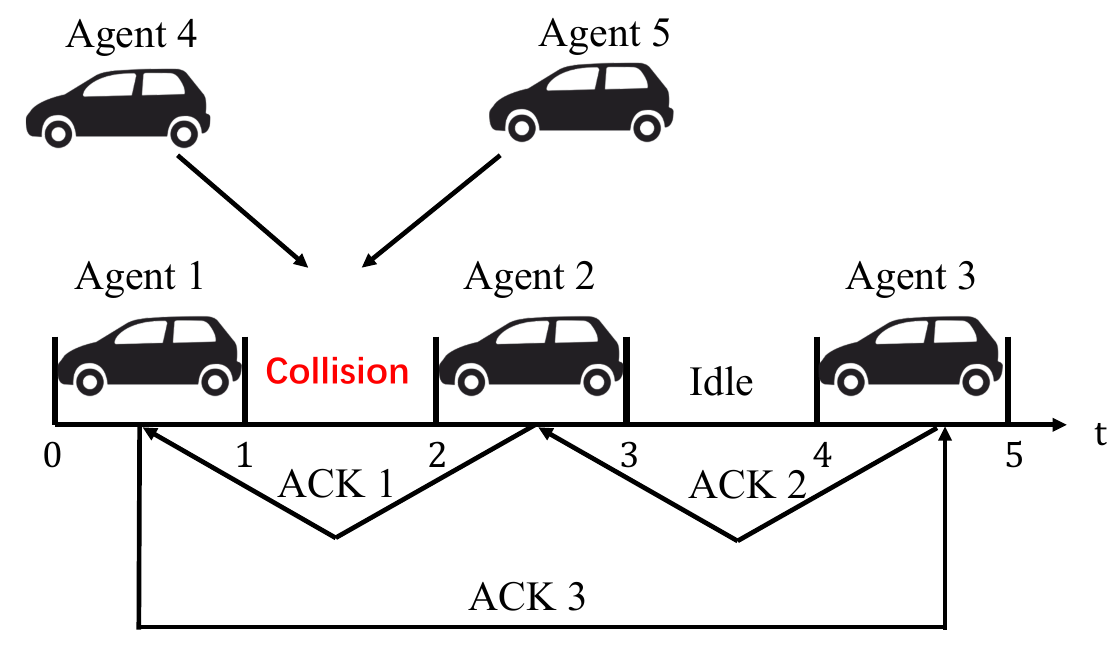}
		}
	\\
	\subfigure[Acks flow when no collision happens.]{
		\includegraphics[scale=0.4]{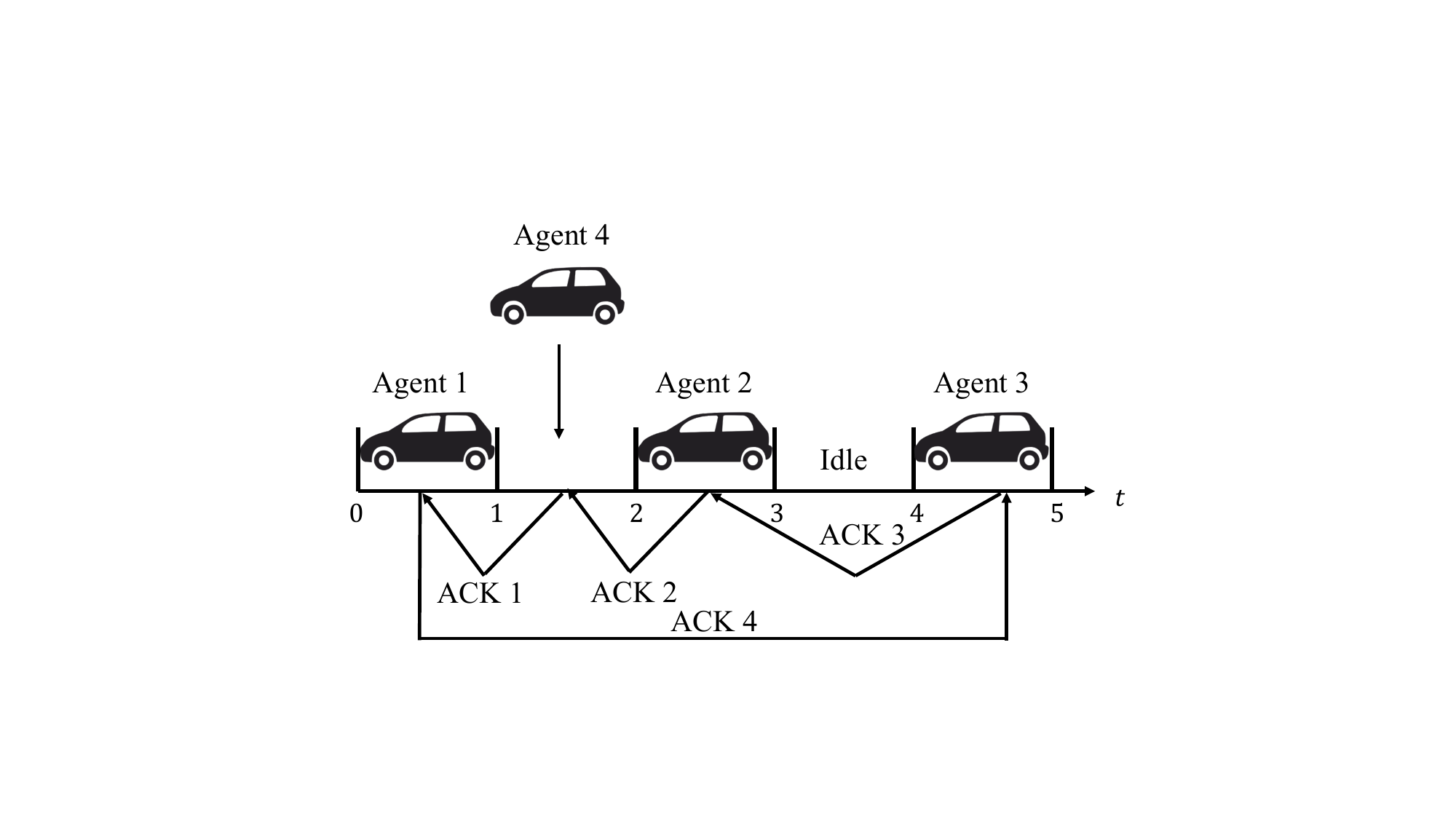}}
	\caption{ACKs flows for the cases with and without collision.}
	\label{fig:ack} 
\end{figure}

However, in general, agents can not support full duplex communications and thus they can not sense the collision when they are sending queries. Therefore, ACKs are needed by these agents to check whether collisions happen. When and only when an agent receives at least one ACK for its query, this agent can confirm that the time slot is taken by itself, otherwise it should continue to randomly select an idle time slot to send its query in following rounds until receiving an ACK. The details of the decentralized multi-access mechanism for queries are described as follows:

\begin{itemize}
\item In every round, agents that have taken their time slots, i.e., have received ACKs in previous rounds, continue to take their time slots to send queries. Other agents sense the idle time slots in the last round and every agent randomly selects one idle time slot (with the same probability) to send query. Notice that if collision happened in a time slot in the last round, this time slot will also be regarded as idle.

\item ACK is a byte attached to the packet of query, which is the index of the last time slot when a query is sent without collision (use 0 if no time slot satisfies this condition). At the end of query phase, the agent that receives the first ACK in the current round will send an ACK, which is the index of the last time slot when a query is sent without collision in the current query phase. Fig. \ref{fig:ack} shows the ACK flows for the cases with and without collision.

\item If an agent receives its ACK, it will take the same time slot in all the following rounds until it leaves the MARL system (even if it does not receive ACKs in the following rounds). Agents that never receive ACKs will randomly select idle time slots to send queries in every round until receiving at least one ACK.
\end{itemize}

With the mechanism above, when more than one agent successfully sends query in a round, all agents that successfully send queries can receive ACKs in this round. The probability that a new agent meets collision when trying to take a time slot is 
\begin{equation}
P_{\rm c} = 1-\left(\frac{T_{\rm idle}-1}{T_{\rm idle}}\right)^{N_{\rm new}-1},
\end{equation}
where $T_{\rm idle}$ is the number of idle time slots in the query phase and $N_{\rm new}$ is the number of agents that have never received ACKs since they entered the MARL system. The assumption that $T_{\rm Q}$ time slots are enough for the transmission of all queries makes sure that $T_{\rm idle} \ge N_{\rm new}$ and thus
\begin{equation}
P_{\rm c} \le 1-\left(\frac{N_{\rm new}-1}{N_{\rm new}}\right)^{N_{\rm new}-1} < 1-\frac{1}{\rm e}.
\end{equation}
This upper bound will only be achieved when $N_{\rm new}$ goes to infinity. If there is not a burst of agents entering the MARL system in the same round, $P_{\rm c}$ can be quite small, even be 0 (e.g. if there is only one new agent entering the MARL in one round).

\textbf{Message phase}: With the queries and message distances broadcast in the query phase, the agents can calculate the message importance of every agent. Although there is no centralized controller in the MARL system, with the calculated message importance, the agents can make the scheduling decisions by themselves. To avoid collision, only agents that send their queries successfully can be scheduled to send messages (because only their message importance can be calculated). 

In the message phase, the agents send messages in order of message importance. As introduced in Sec. \ref{sec:sysmod}, only $N_{\rm c}$ agents that have the highest message importances can send their messages. If the number of agents that send queries successfully is smaller than $N_{\rm c}$, then after these agents finish their transmission, the rest of agents access the wireless channel with CSMA to send their messages until consuming all the time slots in the message phase. 

\begin{figure}[t]
%\vskip 0.2in
%\vspace{-2em}
\begin{center}
\centerline{\includegraphics[width=0.95\columnwidth]{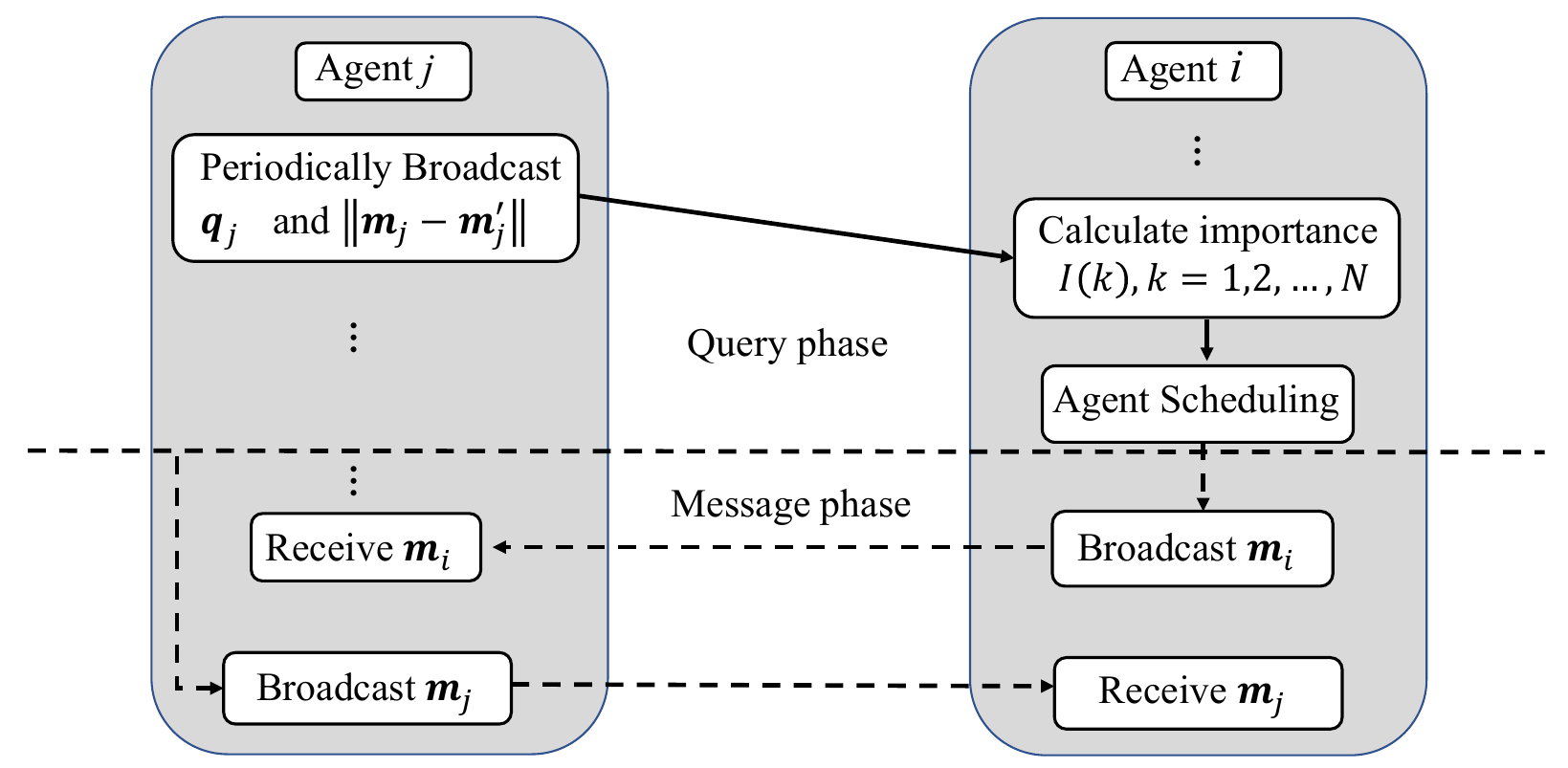}}
\caption{The workflow of communication scheme for decentralized MARL system.}
\label{fig:decflow}
\end{center}
%\vspace{-3.3em}
%\vskip -0.2in
\end{figure}

Fig. \ref{fig:decflow} shows the workflow of the communication scheme for decentralized MARL systems. The key to the proposed importance-aware decentralized multi-access mechanism is that the periodically broadcasted queries help agents realize the message importance of others. Agents share the common knowledge, $\boldsymbol{q}$ and $\|\boldsymbol{m}-\boldsymbol{m}'\|$, with which they can calculate the message importance and generate the same scheduling policy to avoid collision while sending messages.

\section{Message Prediction}\label{sec:predict}

Due to resource limitations, some agents can not be scheduled to share their messages. To perform MARL without these updated messages, which are the input of the learning model, one can exploit history messages or zero vectors as the model input, but this inevitably brings performance degradation. To reduce this performance degradation due to information loss, we propose a message prediction mechanism to provide better model inputs. 

In MARL systems, the messages are vectors generated by neural network $F_{\rm M}$ with observation, encoding the information of agents' states, observations and decisions. Agents need to receive new messages to know the updated information and perform cooperation. If we can predict the updated information according to the old information, then the MARL system can spend fewer communication resources on sharing messages while guaranteeing the system performance. Take the intersection scheduling task as an example, if a vehicle keeps the same driving policy (staying or moving with a constant speed), then we can easily predict its new location, which may be part of its updated message. Note that the prediction only needs to be performed for the messages with relatively small importance, otherwise they should be scheduled to be broadcasted. If an agent has small message importance, its state and decision will be stable with high probability and the prediction of its message will be relatively accurate. This explains why the message prediction can perform well for these agents.

To perform message prediction, we introduce prediction function $F_{\rm p}(\cdot)$, a neural network, into the MARL model. For the training of $F_{\rm p}(\cdot)$, we use the history messages as network input and the newest messages as the groundtruth of the network output. The loss function used for training is $L_2$-norm loss. 

An agent stores every agent's message $\boldsymbol{m}' = \{\boldsymbol{m}'_1,\boldsymbol{m}'_2, \cdots, \boldsymbol{m}'_N\}$, which are broadcasted or predicted, and updates them in every round as shown in Alg. \ref{alg:predict}. Alg. \ref{alg:predict} needs to store every agent's history message and thus the space complexity is $\mathcal{O}(N)$. It needs to predict message for every agent that can not be scheduled to send new message and thus the time complexity is $\mathcal{O}(\max\{N-N_{\rm c},0\})$.

\begin{algorithm}
	
    \caption{Message update with prediction and decision making with predicted messages}
    \label{alg:predict}
    \begin{algorithmic}[1]
    \REQUIRE ~~\\
        Updated messages $ \boldsymbol{m} = \{\boldsymbol{m}_1, \boldsymbol{m}_2, \cdots, \boldsymbol{m}_N\}$;\\
        History messages $\boldsymbol{m}' = \{\boldsymbol{m}'_1, \boldsymbol{m}'_2, \cdots, \boldsymbol{m}'_N\}$;\\
        Message prediction function $F_{\rm p}$;\\
    \ENSURE ~~\\
        Updated history messages $\boldsymbol{m}' = \{\boldsymbol{m}'_1, \boldsymbol{m}'_2, \cdots, \boldsymbol{m}'_N\}$;\\
    Part of agents are scheduled to send updated messages\\
    \FOR{$i=1$ to $N$} 
        \IF{Agent $i$ broadcast its updated message $\boldsymbol{m}_i$}
          \STATE $\boldsymbol{m}'_i = \boldsymbol{m}_i$
        \ELSE
          \STATE $\boldsymbol{m}'_i = F_{\rm p}(\boldsymbol{m}'_i)$
        \ENDIF
    \ENDFOR\\
    Agents use updated $\boldsymbol{m}'$ as RL model input and make decisions
    \end{algorithmic}
\end{algorithm}

After message update (including message broadcast and message prediction) in Alg. \ref{alg:predict}, the updated $\boldsymbol{m}'$ can be used for message aggregation and model input, reducing the performance degradation resulted by missing message sharing. 

Note that the message prediction function does not need extra communication resources and it only uses history (broadcasted or predicted) messages as input. Then why is it able to provide more information for the MARL model input? Firstly, $F_{\rm p}(\cdot)$ has the knowledge of state transition and the agents' policy of making decisions, which is learned from the environment as well as history messages while training and thus adds extra information into the predicted message. Secondly, the message prediction is performed in every round until a new message is received. Therefore, the prediction also utilizes the information of the number of rounds that have passed since the last message broadcast. In other words, the message prediction block helps the MARL model better utilize temporal information to improve the system performance.

\section{Experiment}\label{sec:exp}

\subsection{Experiment Setup}
\begin{figure}[t]
%\vskip 0.2in
%\vspace{-2em}
\begin{center}
\centerline{\includegraphics[width=0.85\columnwidth]{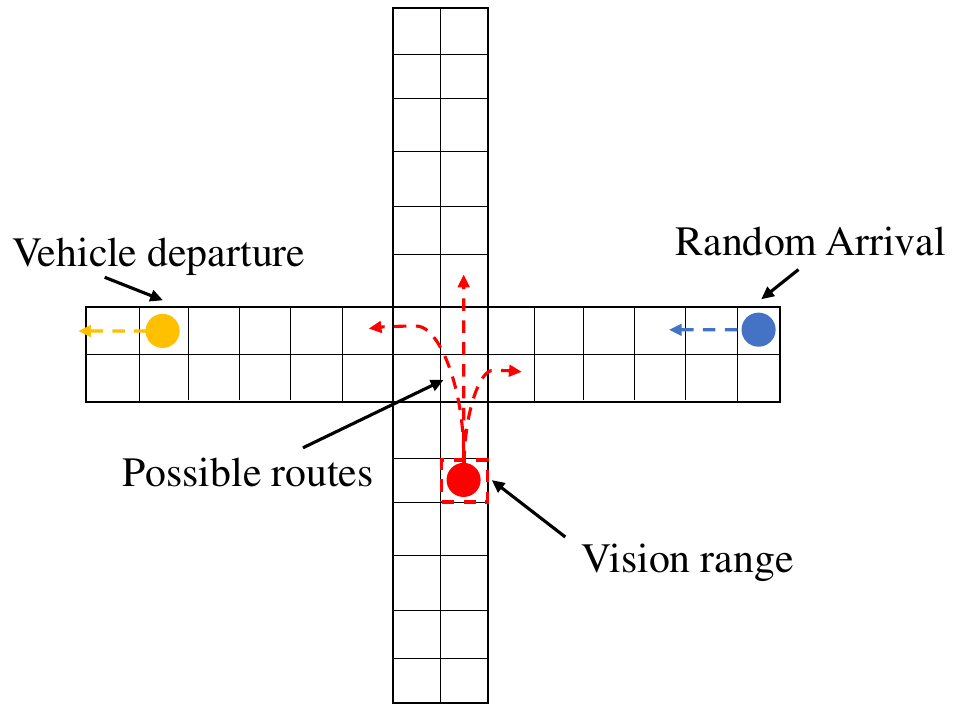}}
\caption{The traffic junction environment in the experiment.}
\label{fig:junction}
\end{center}
\vspace{-2em}
%\vskip -0.2in
\end{figure}

%junction setting
We use a traffic junction environment similar to that of \cite{commnet} to evaluate the performance of the proposed mechanism. Particularly, there is a 4-way junction on a $L \times L$ grid shown in Fig. \ref{fig:junction}. In each round and from each of the four directions, there will be a new vehicle entering the junction with probability $p$. Every vehicle is randomly assigned one of three possible routes to exit the junction as shown in Fig. \ref{fig:junction}. The vehicles in this system do not have a global vision. A vehicle can only observe the state in its vision range, a surrounding $(2V+1) \times (2V+1)$ neighborhood. $V=0$ means the vehicle can only observe the state of the cell of its location.

The actions of a vehicle in any round include: taking one cell forward on its route or staying at its current location. Every vehicle occupies one cell and a collision occurs when more than one vehicle exists in one cell. There are at most $N_{\rm max}$ vehicles in the junction, which means no vehicle will enter the junction if the number of vehicles in the junction reaches $N_{\rm max}$.

In every round, vehicles broadcast their messages via a wireless channel to make better decisions. Due to limited wireless resources, only part of vehicles can broadcast messages in one round. The agents take wireless resources according to the scheduling policy and decentralized multi-access mechanism introduced in Sec. \ref{sec:importance}.

%learning setting
For the input of the MARL model, every vehicle is encoded as a one-hot binary vector set $\{n, l, w\}$, which are its ID, location and assigned route respectively. Because every vehicle can observe an area with $(2V+1) \times (2V+1)$ cells, the state as MARL model input is the concatenation of $(2V+1) \times (2V+1)$ one-hot binary vector sets, whose size is $(2V+1)^2 \times |n| \times |l| \times |w|$. The reward of the MARL system depends on the number of rounds passed since the vehicle enters and if any collision has occurred. The reward of collision is $r_{\rm coll} = -10$ and the reward due to vehicle staying is $\tau r_{\rm time} = -0.05\tau$, where $\tau$ is the number of time slots passed since the vehicle enters the junction. The total reward of the system at time slot $t$ is

\begin{equation}\label{eq:reward} 
r(t) = C(t)r_{\rm coll} + \sum\limits^{N(t)}_{i=1}\tau_i r_{\rm time},
\end{equation}
where $C(t)$ is the number of collisions at time $t$ and $N(t)$ is the number of vehicles in the junction at time $t$. 

In every simulation episode, the MARL system runs $E$ rounds. For simulation and performance evaluation, we define success as no traffic collision occurring in the whole episode. The metric of system performance is average leaving time $\tau_{\rm s}$:

\begin{equation}\label{eq:reward_sum} 
\tau_{\rm s} = \bar{\tau}\mathbb{I}_{\rm s} + E(1-\mathbb{I}_{\rm s}),
\end{equation}
where $\bar{\tau}$ is the average time of leaving the junction of the safely leaving vehicles ($\bar{\tau} = E$ if no vehicle leaves) and $\mathbb{I}_{\rm s}$ means whether the episode achieves success ($\mathbb{I}_{\rm s} = 1$ for success and $\mathbb{I}_{\rm s}=0$ for failure). This metric considers not only the efficiency, but also the safety of the traffic junction system. The simulation results are averaged over $100000$ episodes.  

The parameters used for experiment is summarized in Tab. \ref{tab:param}.

\begin{table}
  
  \begin{center}
    \caption{Summary of experiment parameters}
    \begin{tabular}{cc} % <-- Alignments: 1st column left, 2nd middle and 3rd right, with vertical lines in between
      Parameter & Explanation\\ \hline
      $L$ & The length of the road of the junction \\
      $N_{\rm max}$ & The maximum number of vehicles in the junction \\
      $V$ & Vision of vehicle \\
      $p$ & Arrival rate of vehicle of every direction \\
      $E$ & Number of rounds in one episode\\ \hline
    \end{tabular}
  \end{center}
  \label{tab:param}
\end{table}

In the following simulations, we will use two experiment settings:
\begin{itemize}
\item \textbf{Setting 1}: $L=14$, $N_{\rm max} = 10$, $V=0$, $p = 0.1$, $E=40$.

\item \textbf{Setting 2}: $L=18$, $N_{\rm max} = 15$, $V=1$, $p = 0.1$, $E=60$.
\end{itemize}

The RL model is trained with A3C. We exploit 16 processes to simultaneously train the model with 2000 epochs. The learning rate is 0.001. To get better training performance, the arrival rate of vehicles will first be $p=0.02$ and gradually increase to $p=0.1$ after 1000 epochs.

\subsection{Experiment Results}

%Performance of query and different message size
\subsubsection{Full Communication}
\begin{figure}[t]
%\vspace{-2em}
\begin{center}
\centerline{\includegraphics[width=0.95\columnwidth]{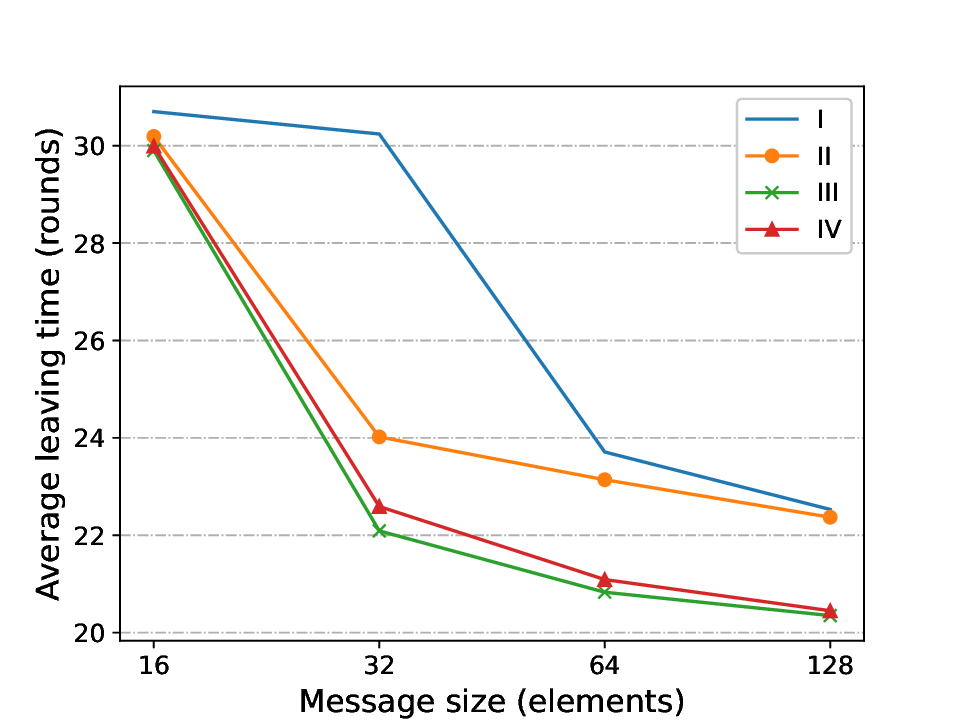}}
\caption{Performance of CommNet (I), CommNet + message block (II), QMNet without query-based message generation (III), and QMNet with query-based message generation (IV) with different message sizes. (Setting 1)}
\label{fig:size1}
\end{center}
%\vspace{-2.5em}
\end{figure}

\begin{figure}[t]
%\vspace{-2em}
\begin{center}
\centerline{\includegraphics[width=0.95\columnwidth]{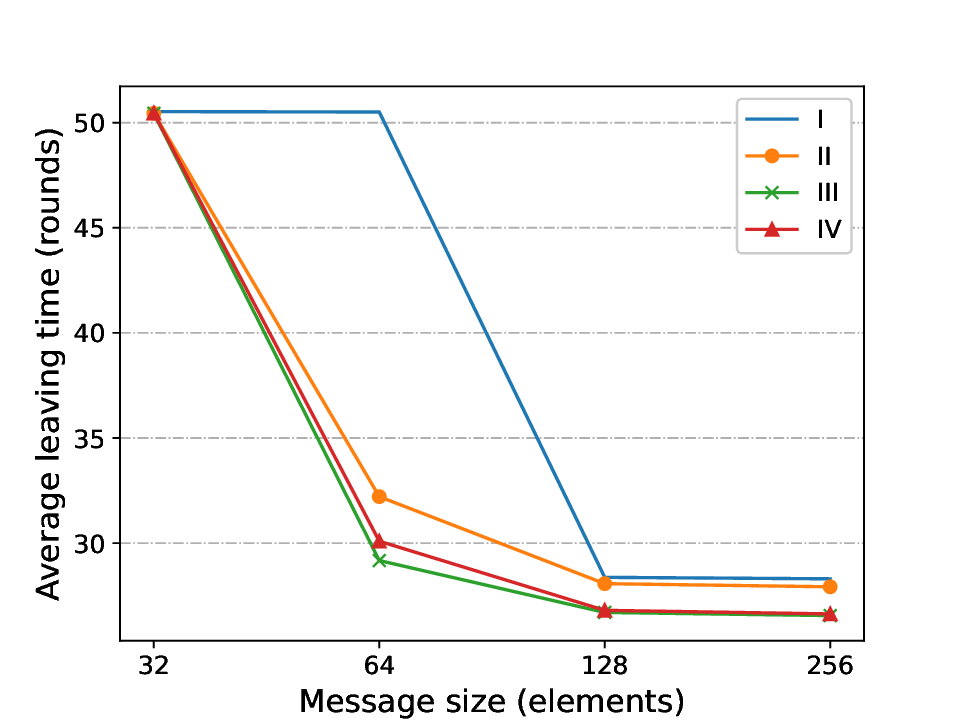}}
\caption{Performance of CommNet (I), CommNet + message block (II), QMNet without query-based message generation (III), and QMNet with query-based message generation (IV) with different message sizes. (Setting 2)}
\label{fig:size2}
\end{center}
%\vspace{-2.5em}
\end{figure}

We first consider the case of full communication, i.e. all agents can broadcast their queries and messages without the constraint of communication resources ($N_{\rm c}=N_{\rm max}$), to evaluate the benefits brought by the proposed message block and query mechanism. The size of query is $s_{\rm Q}=16$. The sizes of hidden units equal the sizes of messages. Fig. \ref{fig:size1} (setting 1) and Fig. \ref{fig:size2} (setting 2) show the performance comparison among CommNet (I) \cite{commnet}, CommNet + message block (II), QMNet without query-based message generation (III), and QMNet with query-based message generation (IV) with different message sizes $s_{\rm M}$. Note that for (IV) the message size in the horizontal axis means the sum of query size and message size, i.e. $s_{\rm Q} + s_{\rm M}$. And for (III) it only means $s_{\rm M}$. 

The message sizes for comparison include $s_{\rm M} = 16, 32, 64, 128$ for setting 1 and $s_{\rm M} = 32, 64, 128, 256$ for setting 2.
It is shown that the performance of CommNet degrades heavily when the message size is decreasing, especially when the message size is reduced to 32 (setting 1) or 64 (setting 2). However, with the proposed message block, the performance has been significantly improved. This is because, with the same message size, the sharing messages processed by the message block can encode more information for cooperation compared with hidden states (which spend part of bits on encoding the information used for the decision making of themselves). Therefore the needed message size to guarantee relatively high performance can be smaller.  
The query mechanism also brings improvement because the weighted aggregation helps the agents filter more important information for decision making.
A significant result is that (IV) achieves almost the same performance as (III) while sending shorter messages. The improvement is obtained by better exploiting the information encoded in the queries. And the messages in (IV) only need to encode the additional information, for which a shorter size is enough.

When the message size reaches 64 (setting 1) or 128 (setting 2), the performance improvement due to increasing message size becomes marginal. Therefore, we choose 64 (setting 1) or 128 (setting 2) as the 'message size' in the following experiments. The size of query is still $s_{\rm Q} = 16$. If not otherwise specified, the QMNet used in the following simulations has query-based message generation block, for which the message size is $s_{\rm M} = 48$ (setting 1) or $s_{\rm M} = 112$ (setting 2)

%Performance of message importance and prediction
\subsubsection{Centralized Scheduling}

Next, we consider the cases with centralized scheduling under different constraints of communication resources. In the following simulations, we do not need to use bandwidth to represent the resource constraint directly, because larger $N_{\rm c}$ (or $T$) can represent larger bandwidth. Because the scheduling is centralized and no communication collision will happen, the constraint of communication resources is represented by $N_{\rm c}$ (instead of $T$), the number of vehicles that can be scheduled to send messages in one round. 

Firstly, we evaluate the performance of the proposed query mechanism with our message importance measure. We compare the performance of QMIX (no communication cost) \cite{rashid2018qmix}, IC3Net \cite{singh2018learning}, SchedNet \cite{kim2019learning} and our proposed QMNet. For QMNet, we use random scheduling and message-importance-based scheduling, including influence-based message importance (the alternative input is zero vector as shown in (\ref{eq:influ})) as well as difference-based message importance (the alternative input is history message as shown in (\ref{eq:diff})). Considering that IC3Net and SchedNet do not need to send query, we set their message sizes to 128 (setting 1) or 256 (setting 2). %waiting for modify 

\begin{figure}[h]
%\vskip 0.2in
%\vspace{-1em}
\begin{center}
\centerline{\includegraphics[width=0.95\columnwidth]{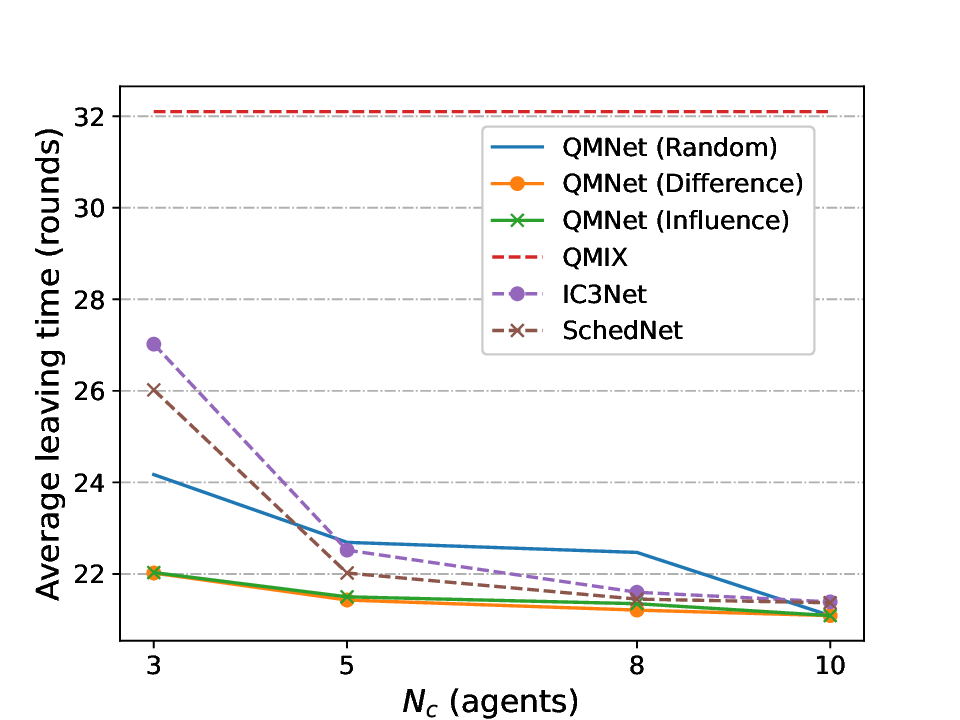}}
\caption{Performance of QMNet compared with other algorithms under different constraints of communication resources (setting 1).}
\label{fig:import1}
\end{center}
%\vspace{-1.5em}
%\vskip -0.2in
\end{figure}

\begin{figure}[h]
%\vskip 0.2in
%\vspace{-1em}
\begin{center}
\centerline{\includegraphics[width=0.95\columnwidth]{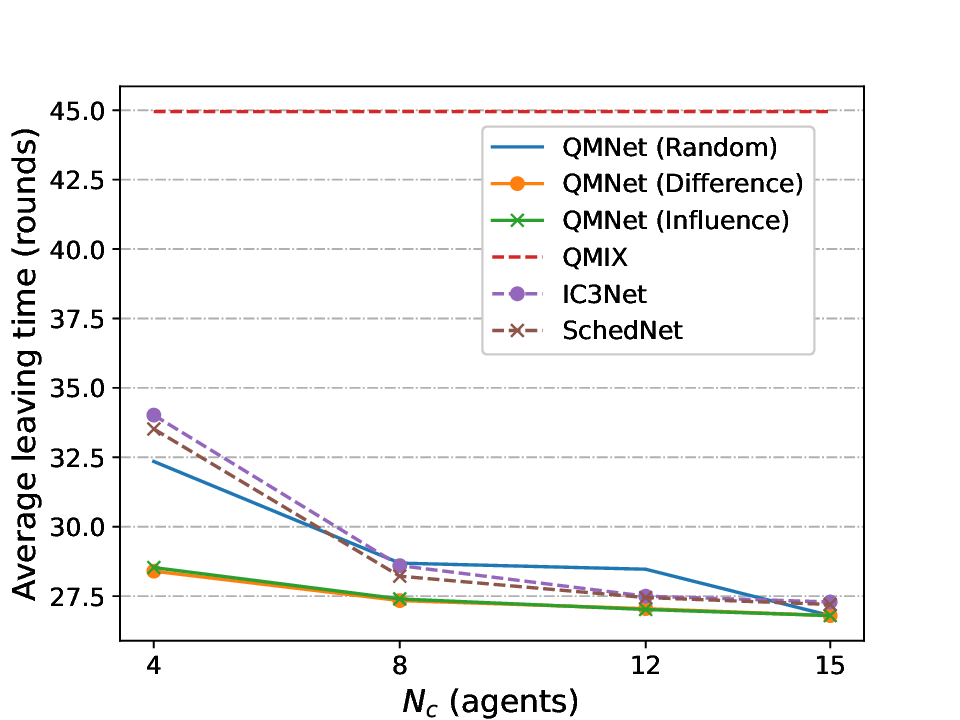}}
\caption{Performance of QMNet compared with other algorithms under different constraints of communication resources (setting 2).}
\label{fig:import2}
\end{center}
%\vspace{-1.5em}
%\vskip -0.2in
\end{figure}

Fig. \ref{fig:import1} (setting 1) and Fig. \ref{fig:import2} (setting 2) show the performance comparison of the algorithms mentioned above under different communication constraints. The communication constraints include $N_{\rm c} = 3, 5, 8$ for setting 1 and $N_{\rm c} = 4, 8, 12$ for setting 2. For comparison, the performances of $N_{\rm c} = 10$ for setting 1 and $N_{\rm c} = 15$ for setting 2, i.e., without constraint of communication resources and all agents can send their messages, are also shown. 
Because QMIX does not need to send messages, its performance does not change while changing communication constraints. The results show that QMIX performs much worse than other communication-based MARL systems. Therefore, communication is important for MARL systems, especially for this traffic junction environment, in which the visions of vehicles are strictly limited.
The performance of these communication-based MARL algorithms shows that the wireless resource limitation results in significant performance degradation. Compared with IC3Net and SchedNet, the proposed QMNet has much less performance degradation. When the number of scheduled vehicles is reduced to 3 (setting 1) or 4 (setting 2), QMNet with importance-aware scheduling has more than 10\% improvement compared with IC3Net and SchedNet. 
In addition, we can find that the importance-aware scheduling can achieve the performance of random scheduling with less than 60\% of the total communication resources. Difference-based message importance performs better than influence-based message importance but the gain is marginal. 

%Fig. \ref{fig:import_p} shows the performance with different $p$, where $N_{\rm c}=5$. It is shown that the gap between random scheduling and the proposed importance-aware scheduling becomes significant when the arrival rate $p$ becomes larger. Therefore, the deployment of communication-efficient scheme is vital for the performance of large-scale MARL system.  

%\begin{figure}[ht]
%\begin{center}
%\centerline{\includegraphics[width=0.9\columnwidth]{import_p}}
%\caption{Performance of query mechanism with different arrival rate $p$.}
%\label{fig:import_p}
%\end{center}
%\end{figure}

\begin{figure}[h]
%\vskip 0.2in
%\vspace{-1em}
\begin{center}
\centerline{\includegraphics[width=0.95\columnwidth]{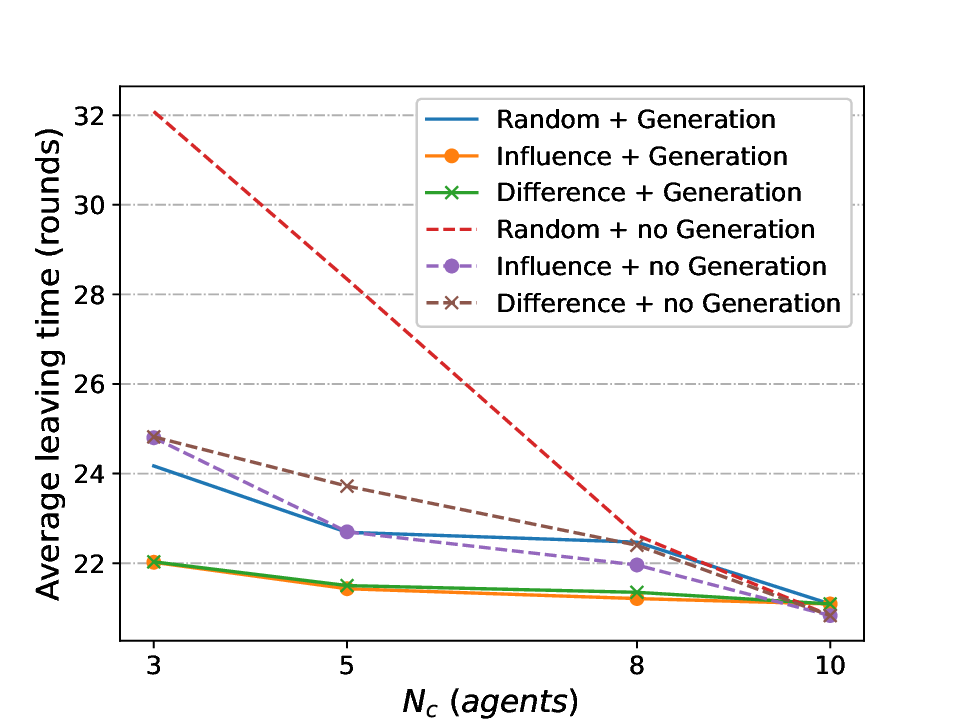}}
\caption{Performance of different scheduling policies with and without query-based message generation under different communication constraints (setting 1).}
\label{fig:aggre1}
\end{center}
%\vspace{-1.5em}
%\vskip -0.2in
\end{figure}

\begin{figure}[h]
%\vskip 0.2in
%\vspace{-1em}
\begin{center}
\centerline{\includegraphics[width=0.95\columnwidth]{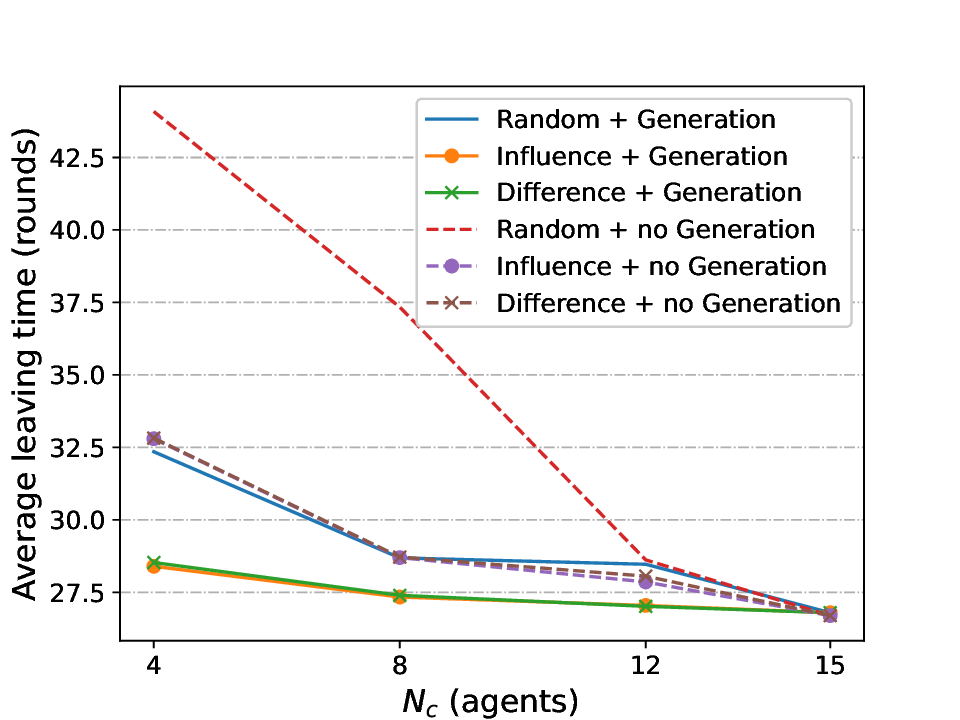}}
\caption{Performance of different scheduling policies with and without query-based message generation under different communication constraints (setting 2).}
\label{fig:aggre2}
\end{center}
%\vspace{-1.5em}
%\vskip -0.2in
\end{figure}

In Fig. \ref{fig:import1} and Fig. \ref{fig:import2}, we can find that even the random scheduling policy can perform better than IC3Net and SchedNet when the wireless resources are scarce. This improvement is credited to the query-based message generation proposed in Sec. \ref{sec:query}. The utilization of the information encoded in the queries helps the vehicles share a significant part of information even when only sending a small part of messages. Fig. \ref{fig:aggre1} (setting 1) and Fig. \ref{fig:aggre2} (setting 2) show the performance of three scheduling policies (random, difference-based message importance, influence-based message importance) with and without query-based message generation under different communication constraints. For the cases without query-based message generation, the message size is $s_{\rm M} = 64$ (setting 1) or $s_{\rm M} = 128$ (setting 2). The performance degradation is significant when query-based message generation block is not applied. Even when exploiting importance-aware scheduling, the system without query-based message generation costs $20\%$ more leaving time when wireless resources are scarce ($N_{\rm c}=3$ for setting 1 and $N_{\rm c}=4$ for setting 2). 

\begin{figure}[t]
%\vskip 0.2in
%\vspace{-2em}
\begin{center}
\centerline{\includegraphics[width=0.95\columnwidth]{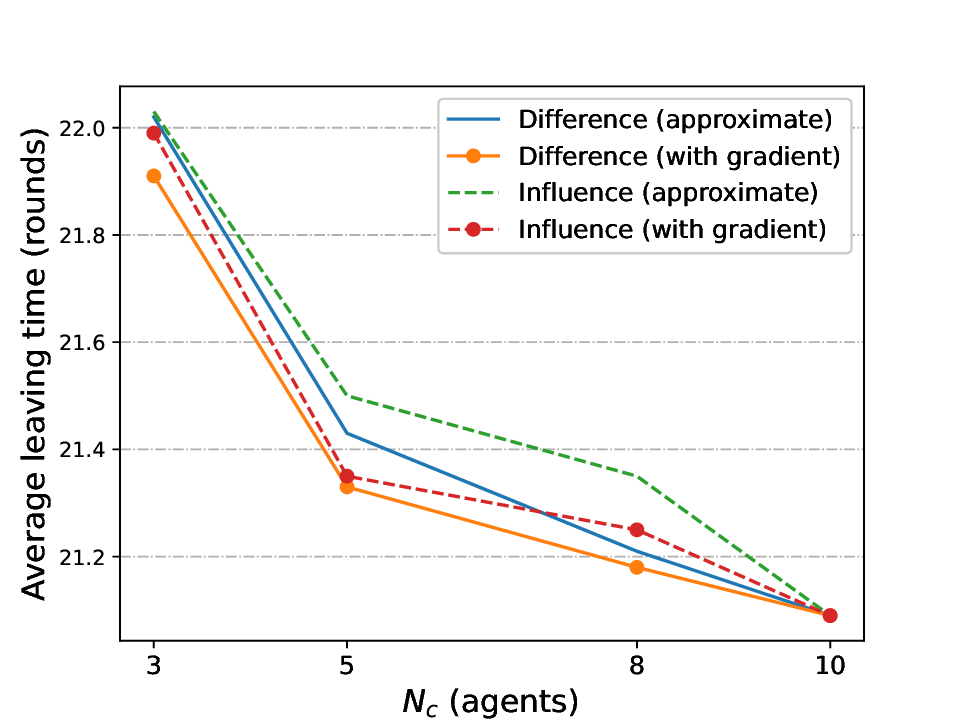}}
\caption{Performance of approximate message importance and the importance using gradients (setting 1).}
\label{fig:grad1}
\end{center}
%\vspace{-2.5em}
%\vskip -0.2in
\end{figure}

\begin{figure}[t]
%\vskip 0.2in
%\vspace{-2em}
\begin{center}
\centerline{\includegraphics[width=0.95\columnwidth]{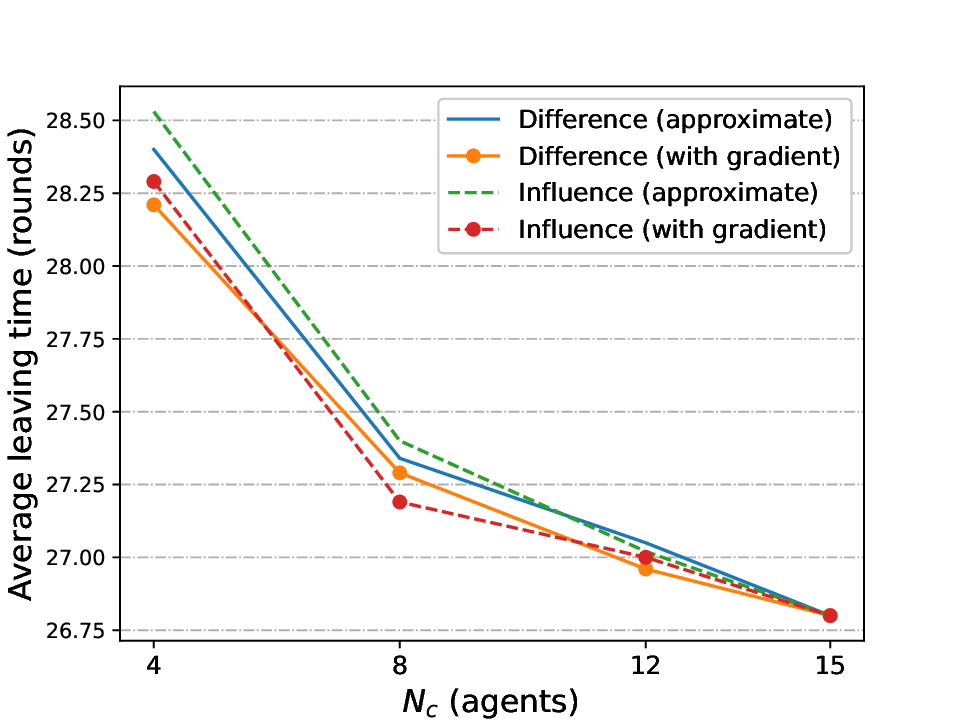}}
\caption{Performance of approximate message importance and the importance using gradients (setting 2).}
\label{fig:grad2}
\end{center}
%\vspace{-2.5em}
%\vskip -0.2in
\end{figure}

In the simplification of message importance in Sec. \ref{sec:importance}, we make some assumptions on the gradients, including uniform distribution of direction and independent distribution of magnitude, to get the approximate message importance. To show whether the assumptions are reasonable, we compare the performance of the derived approximate message importance and the original one using the actual gradients (as shown in (\ref{eq:Ij2})) in Fig. \ref{fig:grad1} (setting 1) and Fig. \ref{fig:grad2} (setting 2). The results show that the scheme with approximated message importance obtains almost the same performance as the one with no approximations. Therefore, it is reasonable to exploit the approximate message importance for scheme design, which needs much less computation and no training. 

\begin{figure}[t]
%\vskip 0.2in
%\vspace{-2em}
\begin{center}
\centerline{\includegraphics[width=0.95\columnwidth]{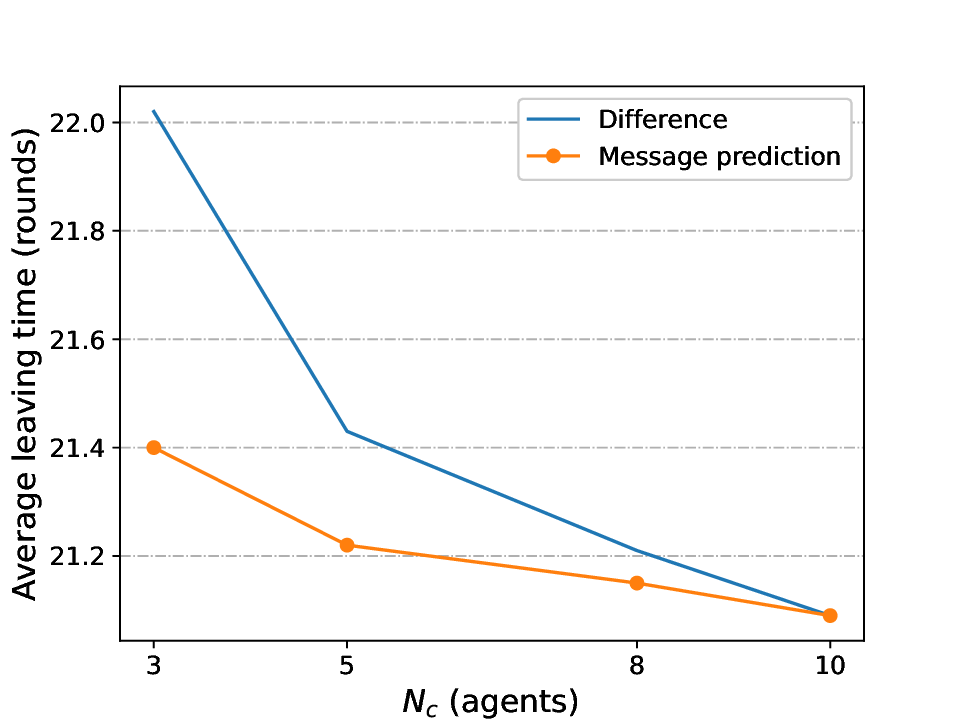}}
\caption{Performance of message prediction under different constraints of communication resources (setting 1).}
\label{fig:predict1}
\end{center}
%\vspace{-2.5em}
%\vskip -0.2in
\end{figure}

\begin{figure}[t]
%\vskip 0.2in
%\vspace{-2em}
\begin{center}
\centerline{\includegraphics[width=0.95\columnwidth]{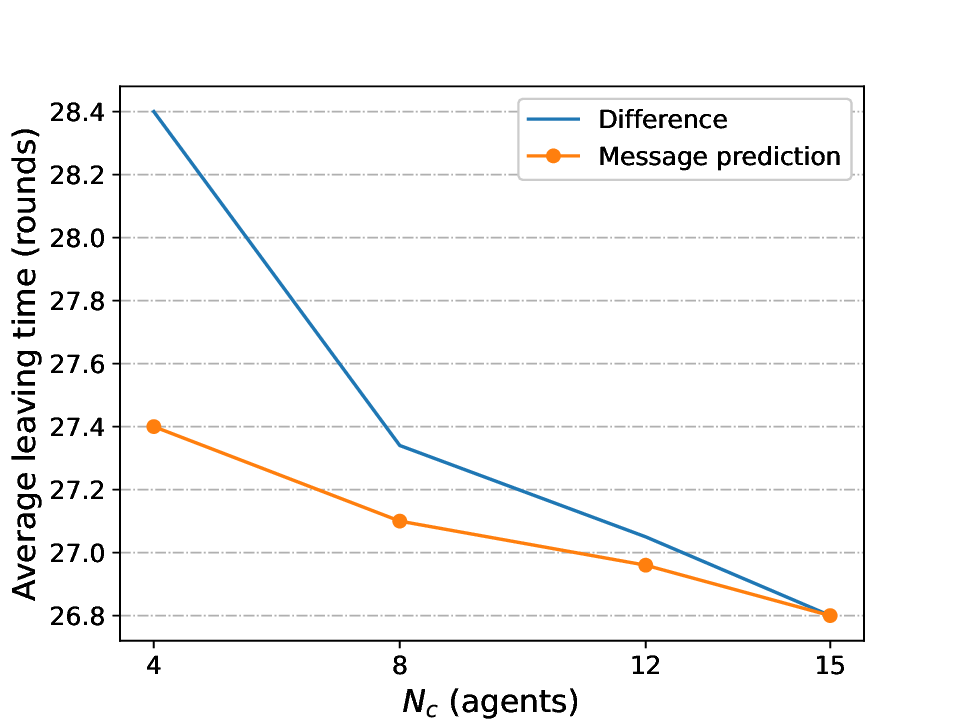}}
\caption{Performance of message prediction under different constraints of communication resources (setting 2).}
\label{fig:predict2}
\end{center}
%\vspace{-2.5em}
%\vskip -0.2in
\end{figure}

Finally, we evaluate the performance of the proposed message prediction. For comparison, we choose a common method that exploits history message (the last shared message) as model input, i.e., the importance-aware message exchange with difference-based message importance. Fig. \ref{fig:predict1} (setting1) and Fig. \ref{fig:predict2} (setting2) shows the performance under different $N_{\rm c}$. With message prediction, the system performance degrades more slowly as the number of scheduled agents becomes smaller. The key to message prediction is providing more information for model input when the agents can not receive new messages. Therefore, with scarce wireless resources, the message prediction can still provide valuable information and obtain substantial performance gain. We can find that with message prediction, the system can obtain the same performance as the one without message prediction, while costing only about $60\%$ of the total communication resources. 

In summary, our proposed schemes, consisting of the query mechanism, importance-aware scheduling and message prediction, can achieve less than $2\%$ performance degradation while saving about $70\%$ of communication resources

\subsubsection{Decentralized Multi-Access}

Finally, we consider the decentralized MARL system and the performance of the proposed importance-aware multi-access scheme. The constraint of communication resources is represented by $T$, the number of time slots in every round. For performance comparison, we exploit CSMA (without priority for agents), IC3Net and SchedNet. For the evaluations of IC3Net and SchedNet, the multi-access mechanisms are based on CSMA and use the generated message importance as the probability that an agent sends messages (contends for the wireless channel) in every time slot. Our proposed importance-aware decentralized multi-access mechanism exploits difference-based message importance to generate scheduling policy (without message prediction, in order to provide a better comparison of the performance of multi-access with other algorithms). 

For QMNet, the settings of message size and query size are the same as those in the simulations of centralized scheduling, for which the costing time slots are $t_{\rm Q} = 1$, and $t_{\rm M} = 3$ (setting 1) or $t_{\rm M} = 7$ (setting 2). The number of time slots in the query phase is $T_{\rm Q} = N_{\rm max} + 1$, i.e., $T_{\rm Q} = 11$ (setting 1) or $T_{\rm Q} = 16$ (setting 2). For general CSMA, the agents with QMNet will send both queries and messages if they access the wireless channel, costing $t_{\rm Q} + t_{\rm M} = 4$ (setting 1) or $t_{\rm Q} + t_{\rm M} = 8$ (setting 2) time slots. For IC3Net and SchedNet, the agents only need to send message, whose size is $s_{\rm M}= 64$ (setting 1) or $s_{\rm M}= 128$ (setting 2), costing $t_{\rm M}= 4$ (setting 1) or $t_{\rm M}= 8$ (setting 2) time slots. The constraints of communication resources are $T=20,26,35,41$ for setting 1 and $T=44,72,100,121$ for setting 2.

\begin{figure}[t]
%\vskip 0.2in
%\vspace{-2em}
\begin{center}
\centerline{\includegraphics[width=0.95\columnwidth]{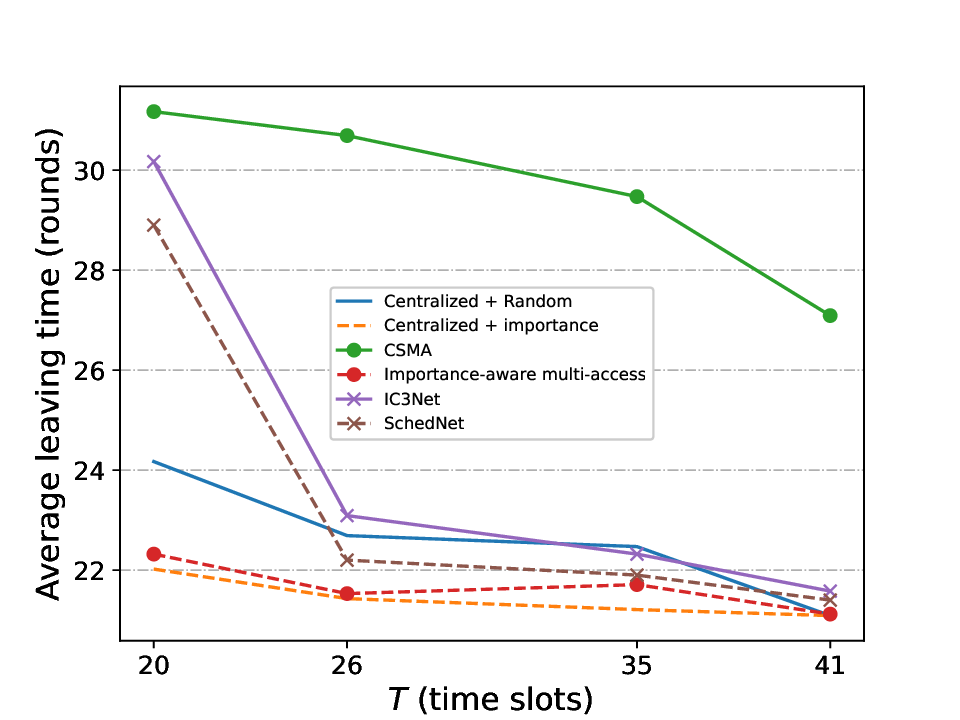}}
\caption{Performance comparison of the decentralized multi-access mechanisms under different constraints of communication resources (setting 1).}
\label{fig:access1}
\end{center}
%\vspace{-2.5em}
%\vskip -0.2in
\end{figure}

\begin{figure}[t]
%\vskip 0.2in
%\vspace{-2em}
\begin{center}
\centerline{\includegraphics[width=0.95\columnwidth]{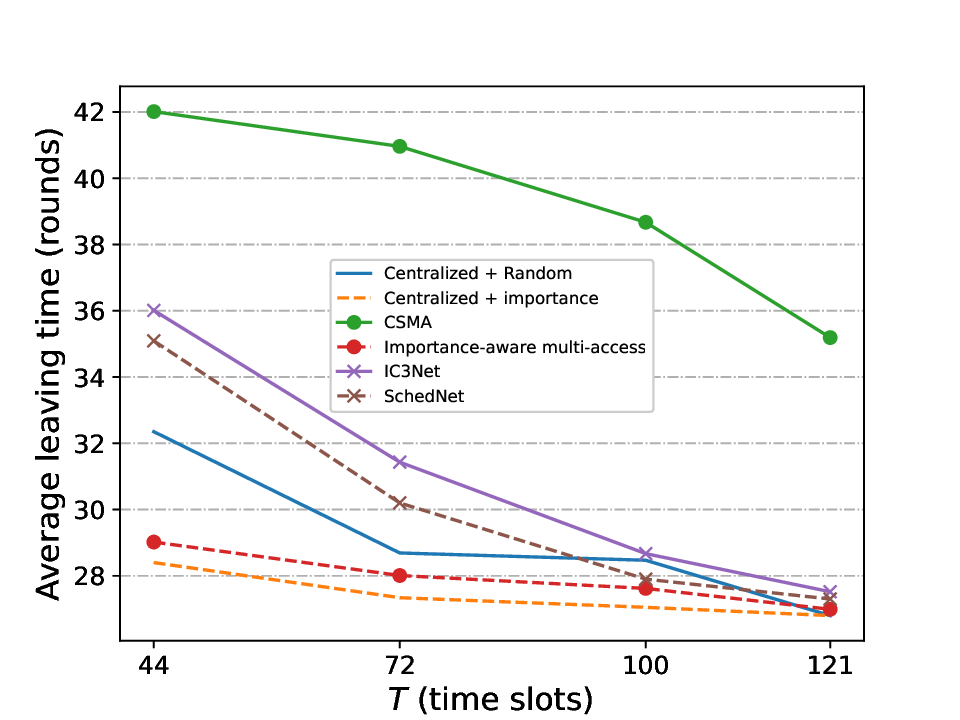}}
\caption{Performance comparison of the decentralized multi-access mechanisms under different constraints of communication resources (setting 2).}
\label{fig:access2}
\end{center}
%\vspace{-2.5em}
%\vskip -0.2in
\end{figure}

Fig. \ref{fig:access1} (setting 1) and Fig. \ref{fig:access2} (setting 2) show the performance comparison of decentralized multi-access mechanisms, where we also show the performance of centralized MARL systems, including random scheduling and importance-aware scheduling (difference-based message importance). Firstly, the performance comparison between general CSMA and centralized scheduling (random scheduling) shows significant performance degradation due to channel contentions. The message-importance-assisted decentralized multi-access mechanisms have much better performance. Our proposed decentralized multi-access mechanism not only outperforms other methods, but also achieves almost the same performance as the centralized case. 

\section{Conclusion}\label{sec:conclusion}

In this paper, we study the MARL systems with limited communication resources by the proposed QMNet with importance-aware communication schemes. The query mechanism helps agents to know others' message importance, then we can select important messages for transmission and design an importance-aware multi-access mechanism to avoid collisions in decentralized MARL systems. To reduce the communication overhead resulted by query mechanism, we propose the query-based message generation block to better utilize information encoded in the query. Besides the message prediction extracts more information from the knowledge of the environment and history messages to help the MARL model perform better.

We evaluate the performance of the proposed schemes in the traffic junction environment under different constraints of communication resources. The results reveal significant performance degradation as the communication resources become limited. Nevertheless, with the proposed QMNet, the system can achieve only $2\%$ of performance degradation while saving about 70\% of communication resources, in centralized MARL systems. For decentralized MARL systems, the proposed importance-aware multi-access mechanism can effectively avoid collisions and achieve almost the same performance as the centralized cases.

Overall, the key of QMNet is costing as few communication resources as possible to share information, with which we can further design schemes to save resources and avoid collisions. Nevertheless, there are still several issues remaining for improvement, including designing robust mechanism to deal with packet loss, developing scheduling policy that considers retransmission, and jointly considering the computation cost of agents. These will be our future research directions.

\bibliographystyle{IEEEtran}
\bibliography{ref}

\end{document}